\documentclass[3p]{elsarticle}
\usepackage{hyperref}

\usepackage{bm}
\usepackage{subcaption}
\usepackage{graphicx}

\usepackage{multirow}
\usepackage{makecell}
\usepackage[table,xcdraw]{xcolor}
\usepackage[export]{adjustbox}
\usepackage{soul}
\usepackage{amsmath}
\usepackage{booktabs}
\usepackage{soul}

\newcommand{\VIVIEN}[1]{\textcolor{black}{#1}}
\newcommand{\YUCHANG}[1]{\textcolor{black}{#1}}
\newcommand{\revise}[1]{\textcolor{black}{#1}}

\biboptions{authoryear}

\usepackage{lipsum}
\makeatletter
\def\ps@pprintTitle{%
 \let\@oddhead\@empty
 \let\@evenhead\@empty
 \def\@oddfoot{}%
 \let\@evenfoot\@oddfoot}
\makeatother

\begin{document}

\begin{frontmatter}

\title{Accuracy and Consistency of Space-based Vegetation Height Maps for Forest Dynamics in Alpine Terrain}

\author[uzh]{Yuchang Jiang}
\corref{ca}
\ead{yuchang.jiang@uzh.ch}
\author[wsl]{Marius Rüetschi}
\author[uzh]{Vivien Sainte Fare Garnot}
\author[wsl]{Mauro Marty}
\author[eth]{Konrad Schindler}
\author[wsl]{Christian Ginzler}
\author[uzh,eth]{Jan D. Wegner}

\cortext[ca]{Corresponding Author}

\address[uzh]{Institute for Computational Science, Universit\"at Z\"urich, Winterthurerstrasse 190,
8057 Zürich,
Switzerland}
\address[eth]{Ecovision Lab, Photogrammetry and Remote Sensing, ETH Z\"urich, Stefano-Franscini-Platz 5,
8093 Zürich,
Switzerland}
\address[wsl]{Department of Land Change Science, Swiss Federal Institute for Forest, Snow and Landscape Research WSL,
Zürcherstrasse 111,
8903 Zürich,
Switzerland
}

\begin{abstract}
Monitoring and understanding forest dynamics is essential for environmental conservation and management. This is why the Swiss National Forest Inventory (NFI) provides countrywide vegetation height maps at a spatial resolution of $0.5~m$. Its long update time of 6 years, however, limits the temporal analysis of forest dynamics. This can be improved by using spaceborne remote sensing and deep learning to generate large-scale vegetation height maps in a cost-effective way. In this paper, we present an in-depth analysis of these methods for operational application in Switzerland. We generate annual, countrywide vegetation height maps at a 10-meter ground sampling distance for the years 2017 to 2020 based on Sentinel-2 satellite imagery. In comparison to previous works, we conduct a large-scale and detailed stratified analysis against a precise Airborne Laser Scanning reference dataset. This stratified analysis reveals a close relationship between the model accuracy and the topology, especially slope and aspect. We assess the potential of deep learning-derived height maps for change detection and find that these maps can indicate changes as small as $250~m^2$. Larger-scale changes caused by a winter storm are detected with an F1-score of 0.77. %
Our results demonstrate that vegetation height maps computed from satellite imagery with deep learning are a valuable, complementary, cost-effective source of evidence to increase the temporal resolution for national forest assessments.

\end{abstract}


\end{frontmatter}

\section{Introduction}

Forest conservation and management requires accurate monitoring of its structure and dynamics. 
Vertical forest structure indicators, such as vegetation height, support biodiversity studies and help forest planning to preserve its functions under increasing ecosystem stress~\citep{dubayah2020global}. Airborne campaigns for vegetation height mapping, while delivering very accurate data, are time-consuming and costly. This leads to a six-year repetition rate of countrywide vegetation height maps in Switzerland, a relatively small territory. 
An alternative to accurate yet costly maps derived from airborne data is vegetation height maps derived from spaceborne data~\citep{lang2019country,potapov2021landsat+gedi,lang2022highResGlobal}. These space-based maps allow for much higher repetition rates at the cost of lower spatial resolution and accuracy. 
In this paper, we present an in-depth analysis of space-based vegetation height maps to explore their practical application in countrywide forest monitoring. 
In particular, we compute countrywide maps over multiple years %
and assess their value for detecting structural change in forests over subsequent years.

A mature, well-established technology to provide detailed and accurate height maps of different types of forests is Airborne Laser Scanning~\citep[ALS;][]{white2016techReviewForestALS}. A common workflow consists in processing the denoised point clouds into a digital terrain model (DTM) using the classified ground points and a digital surface model (DSM) using the highest points. The vegetation height map (VHM) is then computed by subtracting the DTM from the DSM. This standard approach is widely applied in practice in different countries, including Norway~\citep{naesset1997ALS_Norway} and Denmark~\citep{nord2010developingALSDTM}. If a DTM from ALS data exists, additional DSMs can be derived with digital photogrammetry and regular updates are possible~\citep{ginzler2015referencedata}. 
While ALS and photogrammetry certainly do provide high-resolution, accurate data, these techniques come with high operational costs proportional to the area size of the campaign. This ultimately leads to rather low repetition rates. In Switzerland for instance, a still relatively small country of $41,285~km^2$, forests are revisited only every  six years. %

In contrast, Earth observation satellites facilitate more frequent and faster data acquisition at high spatial and temporal resolution.
 ~\citet{piermattei2019pleiades} demonstrate that it is possible to apply photogrammetric methods to Pléiades satellite imagery to produce a 1-meter resolution VHM in Alpine terrain. While showing promising results, a limiting factor of country-scale VHM generation from Pléiades imagery are high costs for buying the satellite images in multi-view stereo configurations. 
Conversely, the Sentinel-2 mission of the European Space Agency (ESA) provides public and free multi-spectral, optical satellite imagery at $10-60~m$ spatial resolution and a revisit time of under six days~\citep{drusch2012sentinel}. 
An alternative approach for VHM generation, inspired by monocular depth estimation in computer vision, is to solely rely on single satellite images without any multi-view configuration.
Although the Sentinel-2 mission does not come with single-pass stereo capability like Pléiades, it is possible to solely rely on single satellite images to generate a VHM, inspired by monocular depth estimation in computer vision~\citep{lang2019country}.
The underlying idea is to directly regress vegetation height per pixel in the Sentinel-2 image based on spectral and textural evidence. 
Because the physical effects that translate vegetation heights to specific spectral values in the satellite images are hard to be modelled directly with sufficient accuracy, data-driven approaches provide a promising alternative. Supervised deep learning approaches are particularly promising.~\citet{lang2019country} propose a convolutional neural network to compute VHM at 10-meter grid spacing from Sentinel-2 satellite images and present countrywide maps for Switzerland and Gabon. \YUCHANG{~\citet{waldeland2022forest} propose a deep learning method to compute a map for the African continent from Sentinel-2 imagery. ~\citet{becker2021country} extend the approach by combining Sentinel-2 images with Sentinel-1 synthetic aperture radar (SAR) images to estimate a range of different forest variables for the entire country of Norway. Although the inclusion of SAR data has shown to enhance model performance, as highlighted by \citet{becker2021country}, their findings have also revealed that optical images from Sentinel-2 play a more crucial role. Furthermore, the exclusion of SAR data only resulted in a minor decrease in accuracy. Given the limited improvement brought by Sentinel-1 in their study and the additional computational requirements, we concentrate solely on utilizing the optical images from Sentinel-2 for this study.} 
Scaling further to global maps of forest structure variables needs globally distributed reference data to train and validate supervised machine learning approaches. The Global Ecosystem Dynamics Investigation \citep[GEDI;][]{gedi2020global} mission, acquiring full-waveform LiDAR data at almost global scale, can provide this kind of reference data.~\citet{potapov2021landsat+gedi} are the first to use GEDI data as reference to train a supervised, bagged regression tree ensemble to produce a global vegetation height map for 2019 with 30-meter resolution. A 10-meter, global VHM of 2020 has been computed by~\citet{lang2022highResGlobal} from Sentinel-2 images and GEDI data. The authors propose a new deep ensemble approach that regresses vegetation heights along with well-calibrated uncertainty estimates per grid cell. These recent methods that densely regress vegetation height maps with supervised machine learning from individual multi-spectral, optical satellite images motivate our work. We investigate to what extent they can be useful in practice to monitor the state of forests at a national scale. 
To that end, we present an in-depth stratified analysis of estimated vegetation heights to understand how the five factors elevation, slope, aspect, forest mix rate, and tree cover density impact performance over multiple years. %
We use two different ALS-based reference datasets of vegetation heights for our analysis: one is used as reference data for training our model whereas the other independent ALS dataset serves as a hold-out dataset to quantify model predictions.

We focus on vegetation height as a forest structure variable in this research because its temporal evolution over multiple years can benefit a wider range of applications in forest dynamics \citep{kugler2015forest}. %
One possible application is tracking structural changes in forests. These changes happen due to anthropogenic forest management practices or natural disturbances such as windthrow, fire, or large-scale bark beetle infestations \citep{sebald21_disturbances}. For example, \citet{honkavaara13_VHMstorm} conducted change detection on remote sensing derived vegetation height maps to spot the disturbed forest area in Finland caused by a storm. %
According to \citet{hall2011multiyear}, analyzing changes in vertical forest structure also supports biomass change estimation over time. We thus believe that computing annual VHMs from individual Sentinel-2 satellite images at country-scale with 10-meter grid spacing has good potential to complement existing airborne campaigns in a meaningful way. 

\YUCHANG{In this paper, we aim to provide a comprehensive and critical evaluation of the capabilities and limitations of deep learning-based vegetation height mapping, for dynamic forest monitoring. We argue that contributing such valuable insights is necessary to advance its practical applications. To achieve this, we set out the following specific objectives:}
\begin{itemize}
    \item 
\YUCHANG{Generate annual countrywide vegetation height maps for Switzerland in the years 2017, 2018, 2019, and 2020, using state-of-the-art deep learning methods inspired by the works of \citet{lang2019country} and \citet{becker2021country}.}
    \item 
\YUCHANG{Thoroughly validate the accuracy of these vegetation height maps against an independent ALS dataset of vegetation heights.}
    \item 
\YUCHANG{Evaluate the temporal and spatial generalization ability of our model on unseen years and regions because understanding the generalizability of the model is important in cases where aerial data are not available for a specific year or region.}
    \item 
\YUCHANG{Analyze how our vegetation height maps can effectively support the detection of annual forest structural changes, and identify the limitations concerning the size of the area for which these maps can provide accurate results.}
\end{itemize}

\section{Materials and methods}

\subsection{Study area}

Our study covers Switzerland, a mountainous country with a total surface area of $41,285~km^2$. Switzerland's territory is composed of $31.3\%$ forest and woods, $23.4\%$ agricultural, $12.4\%$ alpine farmland and the remaining are urban or unproductive areas \citep{FSO2021}. According to the fourth survey of the NFI \citep{NFI4_2020}, the forest in Switzerland includes: 41\% pure coniferous forests, 19\% mixed coniferous forests, 14\% mixed deciduous forests, and 24\% pure deciduous forests. \revise{Almost two thirds of the forest area in Switzerland is regularly managed. The management is small-scale, large-scale clear-cutting does not occur. However, natural events such as storms or rare fires can lead to large-scale damage. The proportion of old-growth forests is high by European standards.} The southern part of Switzerland is mostly occupied by the Alps, while the northern side has many urban areas on the Swiss plateau. %
The elevation of Switzerland ranges from 193 meters to 4634 meters above sea level. This wide range of elevation, topography, and the diversity of forest cover types make Switzerland a challenging study area, suitable for an in-depth analysis of the model predictions.

\subsection{Data}

\subsubsection{Reference data}

\begin{figure}
	\centering
		\includegraphics[width=.7\textwidth]{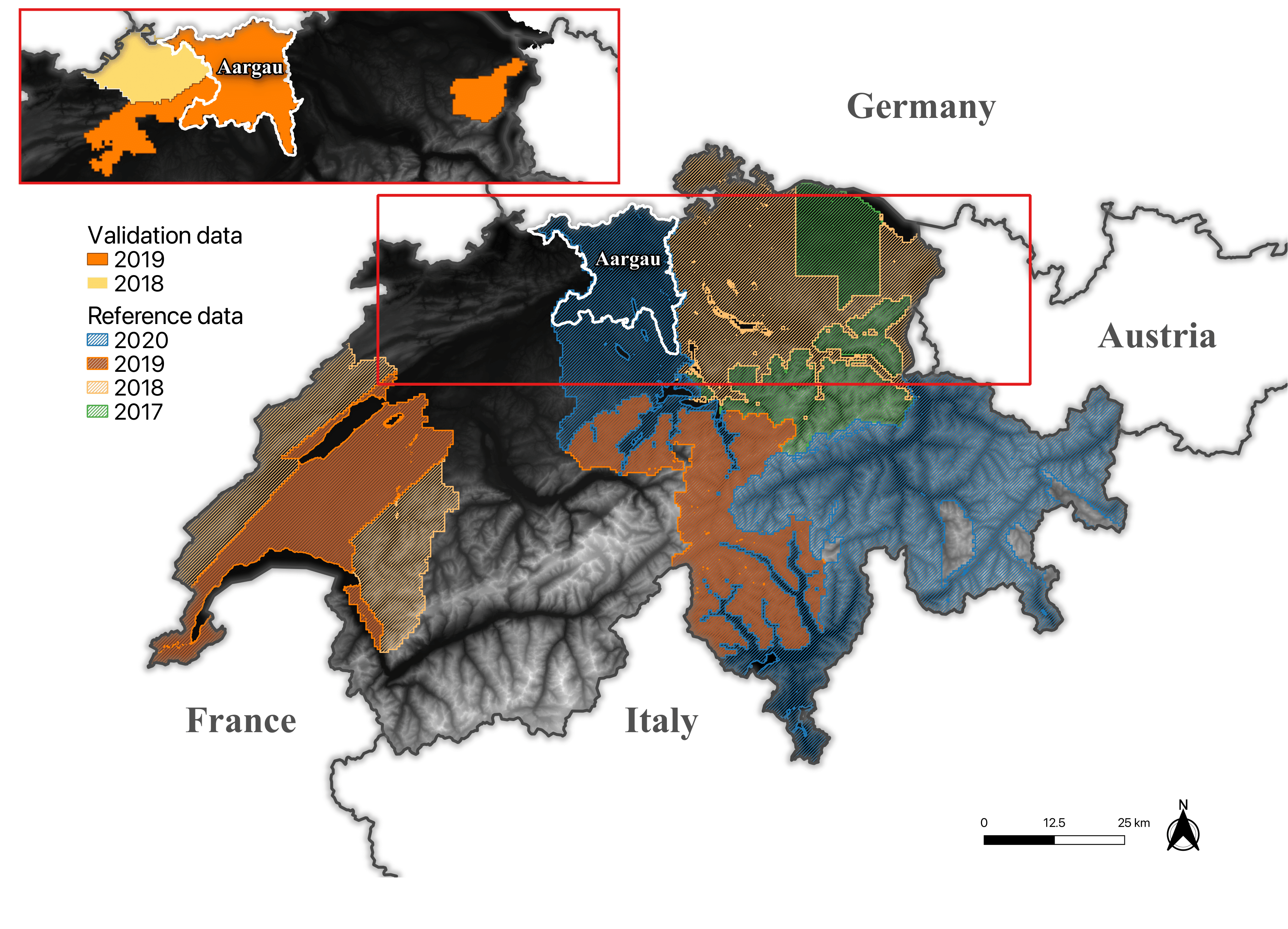}
	\caption{Overview of the reference data ($VHM_{ALS}$) and validation data ($VHM_{V}$). The reference data is based on the acquisition of ALS data in the years 2017 to 2020. The validation data shown in the red rectangle is the cantonal ALS data captured in 2018 and 2019. The DTM of Switzerland is shown as the background.}
	\label{FIG:reference_year_ch}
\end{figure}

\begin{figure}
     \centering
     \begin{subfigure}[b]{0.7\textwidth}
         \centering
         \includegraphics[width=\textwidth]{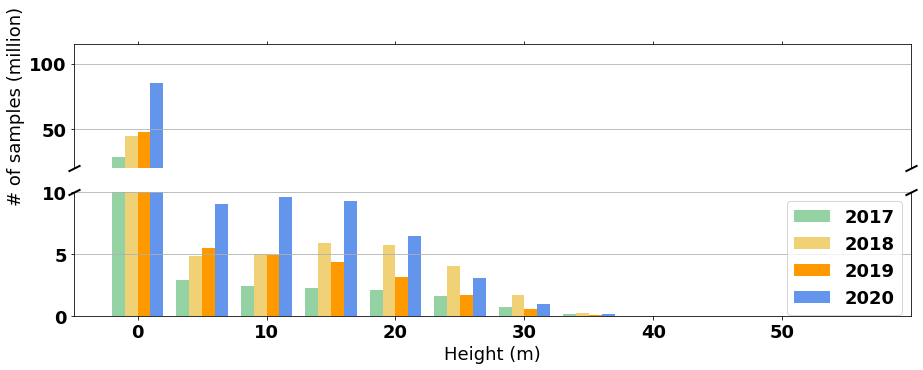}
         \caption{Distribution of mean vegetation height}
         \label{fig:dist_mean}
     \end{subfigure}
     \begin{subfigure}[b]{0.7\textwidth}
         \centering
         \includegraphics[width=\textwidth]{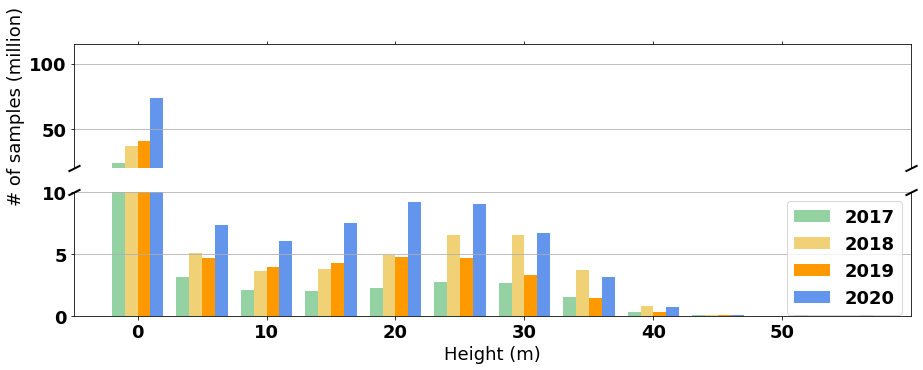}
         \caption{Distribution of max vegetation height}
         \label{fig:dist_max}
     \end{subfigure}
        \caption{Distribution of mean and maximum vegetation heights over different years. Note that we are dealing with a highly imbalanced distribution of vegetation heights.}
        \label{FIG:dist_reference_data}
\end{figure}

We use an ALS-based reference dataset for model training and testing. Data were acquired as part of the ongoing national ALS campaign swissSURFACE3D \revise{with a mean point density of $15-20~pts/m^2$} \citep{swisssurface3d}. \revise{The data are delivered as point clouds semantically classified using the TerraScan software and checked by visual interpretation.} The data used in this study cover the years 2017-2020 for 66.5\% of Switzerland (Figure~\ref{FIG:reference_year_ch}). 
To derive heights above ground, the point clouds are normalised using the information of the ground classified points (ASPRS class 2 \citep{ASPRSLASspecification}). Then only points classified as vegetation (ASPRS classes 3, 4, 5) by swissSURFACE3D are chosen for the generation of the map containing the vegetation height with a pixel spacing of $1~m$. As the last step, we resample this reference data of 1-meter resolution, to match the 10-meter ground sampling distance (GSD) of Sentinel-2 images. We use average and maximum pooling to produce two different outputs ($VHM_{ALS}$): the vegetation height mean and the vegetation height maximum value in the \revise{$100~m^2$} pixel. 
The distributions of the resampled mean and maximum vegetation heights are shown in Figure \ref{FIG:dist_reference_data}, revealing an inherent imbalance in our data: a considerable number of samples correspond to low vegetation heights.

\subsubsection{Input data for VHM estimation}
\begin{figure}
	\centering
		\includegraphics[width=.7\textwidth]{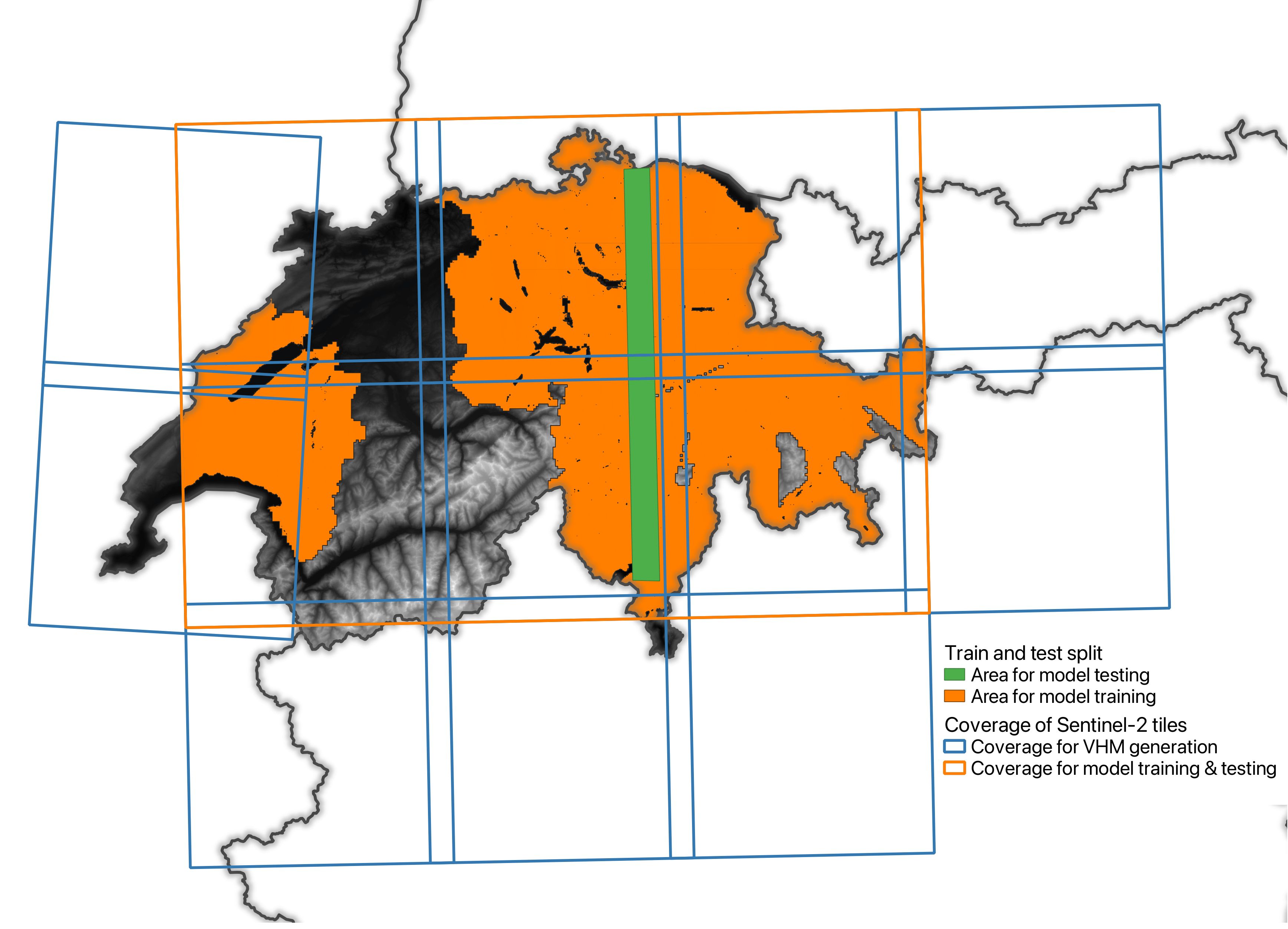}
	\caption{Overview of the Sentinel-2 tiles and reference data used for training and testing. Although we need 13 Sentinel-2 tiles (blue tiles) to cover Switzerland completely, six Sentinel-2 tiles are enough to cover most of the country. To reduce the computational burden, we use the areas covered by those six tiles for model training and testing (areas inside the orange rectangle). A north-south transect of $1800~km^2$ is defined as a hold-out test area unseen by the model during training.
}
	\label{FIG:train_test_split}
\end{figure}

We use the Sentinel-2 level-2A product~\citep{sentinel2}, which provides bottom-of-atmosphere land surface reflectances corrected for atmospheric effects. Each Sentinel-2 tile covers a $100 \times 100~km^2$ area and there are multiple observations per tile per year due to a revisit cycle of 3-5 days. We use those six tiles that cover $96\%$ of the Swiss territory for model training and testing, and 13 tiles to generate complete countrywide maps, as shown in Figure \ref{FIG:train_test_split}. We refer the VHM predicted by our model to $VHM_{S2}$.

From all available Sentinel-2 images of a year, we select those acquired during the leaf-on season from May to September. We use the cloud probability mask of the Level-2A product to select the ten observations with the lowest cloud probability per year per tile. We use the four spectral channels with 10-meter spatial resolution: red, green, blue, and near-infrared as they provide high spatial resolution and have been demonstrated to be effective for vegetation height estimation in \citet{lang2019country}.

We also make use of the swissALTI3D DTM provided by swisstopo \citep{swissalti3d} as additional input data to the model. For completeness, we note that swissALTI3D is generated from ALS data and resampled into 1-meter pixels. To match the Sentinel-2 GSD, swissALTI3D is resampled to a 10-meter grid by averaging across $10\times10$ meter neighborhoods. 

\subsubsection{Validation data}
For the evaluation of the predicted $VHM_{S2}$, we use an independent ALS-based validation dataset. The data was acquired by six Swiss cantons in the years 2018 and 2019 \revise{with a mean point density of $25-30~pts/m^2$} \citep{CantonAG_ALS,CantonBL_ALS,CantonAR_ALS,CantonSO_ALS}. The overview of the areal coverage of this cantonal validation data is shown in Figure \ref{FIG:reference_year_ch}. These cantonal ALS datasets are processed and then resampled into a $VHM_{V}$ with a pixel spacing of $10~m$ in the same way as the swissSURFACE3D. \revise{We deem the slightly higher mean point density of the cantonal validation data as neglectable since the vegetation height information was aggregated to a pixel spacing of $10~m$ in both cases}.
\YUCHANG{Coincidentally, in the case of the Canton of Aargau, the cantonal data $VHM_V$ covers the region in 2019, and the same area was also captured by the national swissSURFACE3D campaign in 2020.}
 The correlation analysis between the ALS data of Aargau, acquired by the canton in 2019, and the national swissSURFACE3D campaign reveals a robust correlation with an $R^2$ of 0.92, demonstrating consistent alignment between our reference and independent validation datasets. \YUCHANG{Therefore,} the ALS data of Aargau captured in 2019 and 2020 enables an analysis concerning the sensitivity of $VHM_{S2}$ to detect structural changes in the forest of the Canton of Aargau with a forest area of approximately 500 $km^{2}$. A comparison between the biannual differences captured by the ALS and $VHM_{S2}$ is possible. In these two years, the ALS survey was conducted in spring, from March to April in 2019 and from February to March in 2020. Consequently, the time period is not exactly the same as that of $VHM_{S2}$. Most structural changes should, nevertheless, appear in both $VHM_{S2}$ and ALS data. Forestry interventions occur mainly in winter and major disturbances like windthrows were not reported for the region between March and the leaf-on season 2020: we assume that the different acquisition times of the ALS data and $VHM_{S2}$ can be ignored. 

We use additional countrywide products to mask out non-forested areas for the analysis and define different strata to thoroughly analyse the accuracy of the $VHM_{S2}$ in 2018 and 2019. For masking, we use the forest mask of the Swiss NFI \citep{waser2015_forestmask}. Raster products of topography, forest mix rate \revise{regarding the share of broad-leaved vs. coniferous trees}, and tree cover density are used for the strata analysis. We compute information on elevation, slope and aspect of the terrain from swissALTI3D and obtain the forest mix rate product from the NFI \citep{waser2021_mixrate}, and the tree cover density from the Copernicus High Resolution Layer 2018 \citep{Copernicus_HRLTCD}. To achieve a consistent spatial resolution for all the data in the analysis, all raster products are resampled to the $VHM_{S2}$ grid of 10 meters with bilinear interpolation.

As it is common to encounter new or updated data in vegetation height mapping applications, utilizing an independent dataset allows for the evaluation of the model's performance on previously unseen data, providing a crucial assessment of its ability to handle new information. In addition to assessing generalization, this independent dataset provides a substantial amount of data for stratified analysis and assists in evaluating the detection of structural changes.

\subsection{Methods}
\begin{figure}[!htbp] 
	\centering
		\includegraphics[width=.8\textwidth]{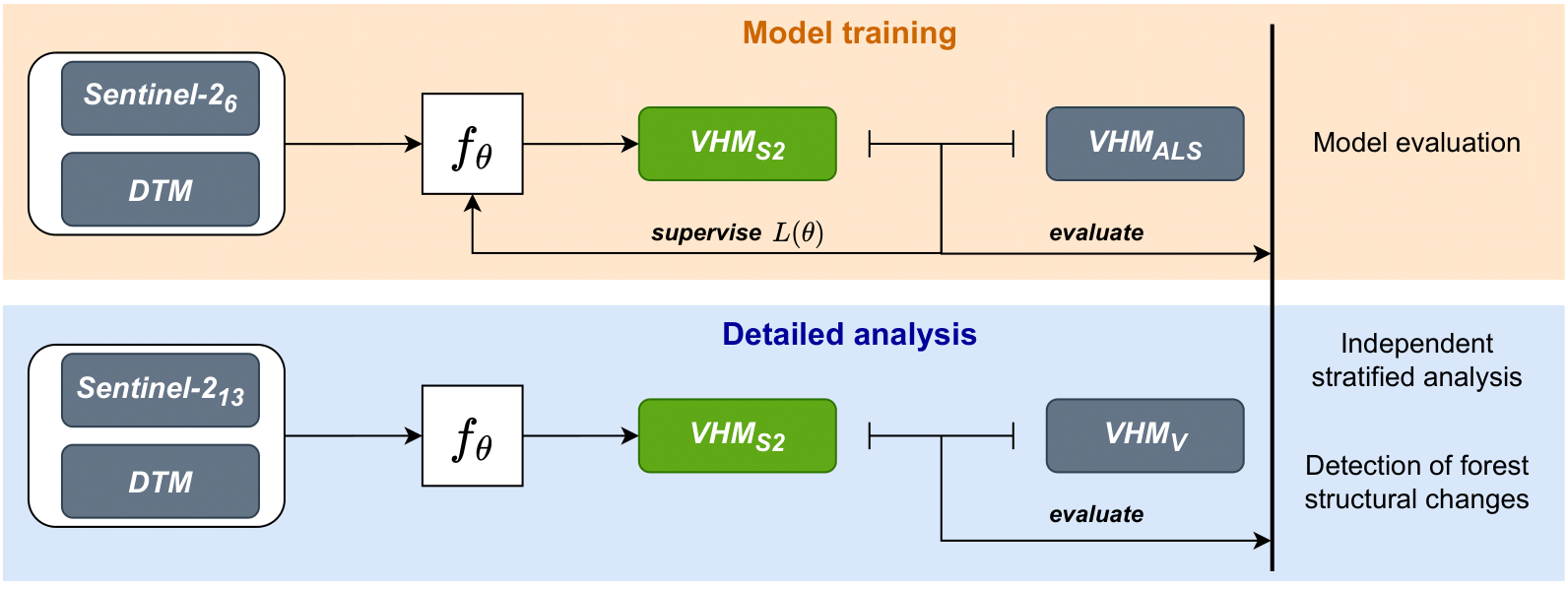}
	\caption{Overview of our study design: (1) train and test the model based on reference data $VHM_{ALS}$ and six Sentinel-2 (S2) tiles, and (2) use the trained model to generate countrywide map $VHM_{S2}$ on 13 S2 tiles; independent analysis based on validation data $VHM_{V}$.}
	\label{FIG:flowchart}
\end{figure}

\YUCHANG{We show our overall experimental setup in Figure  \ref{FIG:flowchart}.} We first train and test the model on the reference data $VHM_{ALS}$. We then generate complete countrywide maps and perform a detailed analysis based on the independent validation data $VHM_{V}$.

\subsubsection{Model architecture}

\begin{figure}[!htbp] 
	\centering
		\includegraphics[scale=.5]{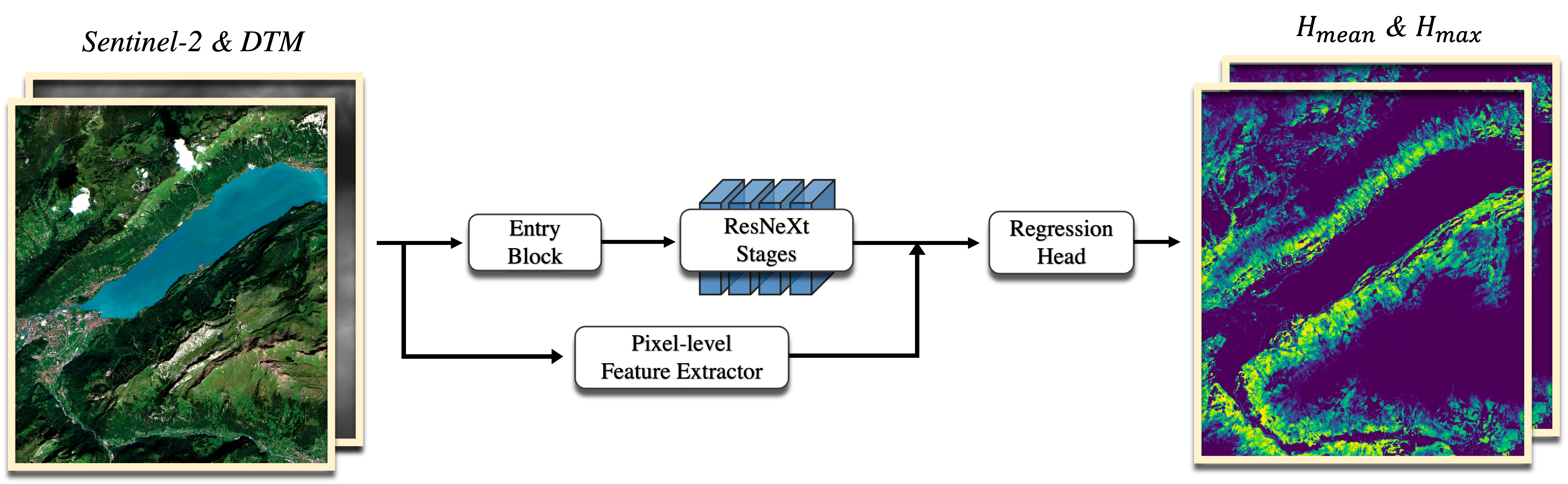}
	\caption{Our model combines a deep feature extractor based on the ResNeXt backbone with a pixel-level feature extractor to account for low-level and high-level features. The two feature maps are concatenated and processed by the regression head to predict the mean and max vegetation height.}
	\label{FIG:resnext_model}
\end{figure}

We use a fully convolutional neural network based on the previous work of \cite{becker2021country} and the ResNeXt architecture \citep{xie2017resNeXt} for our study. We do, however, modify the input and output layers of the model architecture proposed in \cite{becker2021country} to adapt to our settings (but keep all other parts unchanged): we remove the input layers designed for Synthetic Aperture Radar data and also the output layers for uncertainty estimation. As shown in Figure \ref{FIG:resnext_model}, our model consists of an entry block, four ResNeXt stages, a low-level feature extractor, and a regression head to generate the final output. The Sentinel-2 input and the elevation layer are concatenated and given as input to the model. 
The entry block processes the combined input of the satellite image and DTM. It consists of one 1x1 convolution layer, batch normalization, and applies ReLU as an activation function. \YUCHANG{Four stages of ResNeXt modules that extract high-level features follow this entry block. We provide a detailed description of the ResNeXt stage in Figure \ref{fig:resnext_stage_arch} in the Appendix.}
Another feature extractor then extracts low-level image features from raw pixels. It consists of three layers: two 1x1 convolutional layers with a single activation layer (ReLU) in the middle. After concatenating low-level and high-level features, the regression head produces the final refined mean and maximum height maps. It \YUCHANG{comprises one convolutional layer, a ReLU activation function, and another convolutional layer.} 

\subsubsection{Model training}

We frame our task as a mono-temporal, pixel-based regression problem. Our model $f_{\theta}$ regresses the vegetation height value $y_i$ of a 10-meter pixel from an input patch $x_i$ of shape $H\times W$. The input patch $x_i$ contains one Sentinel-2 observation concatenated with the DTM.
We empirically choose the mean absolute error (MAE) loss with L2-penalty to supervise the model's parameters $\theta$ :
\begin{equation}
    L(\bm{\theta}) = \frac{1}{N} \sum^N_{i=1} |f_{\bm{\theta}}(x_i) - y_i| + \lambda \|\bm{\theta}\||^2_2 \;\; ,
\end{equation}
where $N$ is the number of training samples and $\lambda$ is the hyperparameter of weight decay.

We split the reference data into two geographically separate areas for training and testing. During training, we decompose the $100 \times 100~km^2$ Sentinel-2 image into patches of size $15 \times 15$ pixels (i.e., $22,500~m^2$ area per patch on the ground) following \citet{lang2019country}. We select those patches where valid target $y_i$ exists for the center pixel and the cloud probability of the center pixel is under $10\%$ to avoid noisy training samples. We refer to these patches as \emph{valid} patches. \YUCHANG{We consider different valid patches of the same location as distinct samples during training and we supervise all images of the same location in one given year with the same target value to neglect intra-annual changes.} Although using all images of the same location helps to reduce the atmospheric effect introduced by individual observations, it also brings redundant information and slows down the model training process. As a compromise between robustness and efficiency, we rank the images by the number of valid patches and select two Sentinel-2 images of the same location for each year.

\subsubsection{Implementation details}
We normalize per channel for all input images for training and testing datasets based on statistics computed solely from the training dataset. During training, we randomly sample a subset of patches ($64,000$ patches) among the dataset in each epoch due to the large dataset ($869,183$ patches). We implement our model in PyTorch \citep{paszke2017automatic_pytorch} and train with $batch\_size = 64$, $learning\_rate = 10^{-5}$. For the ResNeXt stages in the model architecture, we define $groups = 32$, $width\_per\_group = 4$, $N_{block}$ for the four stages $[2, 3, 5, 3]$, following \citet{becker2021country}. Here $groups$ and $width\_per\_group$ are the hyperparameter of grouped convolutions inside the ResNeXt block and $N_{block}$ defines the number of ResNeXt blocks inside each ResNeXt stage. \YUCHANG{We show the detailed configuration of the model in Table \ref{tab:detailed_model_arch} in the Appendix.} We apply Adam \citep{kingma2015adam_iclr} for optimization with parameters $\beta_1 = 0.9$, $\beta_2 = 0.999$ and $\epsilon = 10^{-8}$. We further split the training dataset into training and validation sets with an 80:20 ratio to monitor the model training and to avoid over-fitting. \YUCHANG{We train the model for 500,000 iterations until it convergences on the validation set}, which takes approximately 50 hours on a single NVIDIA GeForce RTX 2080 Ti GPU.

\subsubsection{Evaluation of vegetation height estimation}
We first evaluate our model's predictions against the test set obtained from the same source of reference data ($VHM_{ALS}$). For inference, we normalise the test set using the normalisation values of the train set. We split each image into patches of $512 \times 512$ pixels with 16 pixels overlap to produce a large-scale map efficiently. After model prediction, we recompose patches into the original image size and mask out cloudy (cloud probability $> 10\%$) and non-vegetated (classified as water or snow according to Sentinel-2 land cover classification) pixels. Lastly, we take the median of the predictions obtained from different images of the same location in a given year to produce the final annual map. We thus neglect the height variations within a single year, which are outside of the scope of our study.

We evaluate our model's performance based on the following metrics: mean absolute error (MAE), root mean squared error (RMSE), and mean bias error (MBE). Although we can assess the performance with MAE alone, the RMSE can make outlier analysis more transparent. The MBE is important to detect biases in model predictions like underestimation and overestimation.

\begin{equation}
    MAE = \frac{1}{N} \sum^{N}_{i=1} | f_{\bm{\theta}}(x_i) - y_i |
\end{equation}
\begin{equation}
    RMSE = \sqrt{\frac{1}{N} \sum^{N}_{i=1}  (f_{\bm{\theta}}(x_i) - y_i)^2}
\end{equation}
\begin{equation}
    MBE = \frac{1}{N} \sum^{N}_{i=1} f_{\bm{\theta}}(x_i) - y_i 
\end{equation}

\begin{equation}
    MAEr = \frac{\frac{1}{N_{s}} \sum^{N_{s}}_{i=1} f_{\bm{\theta}}(x_i) - y_i}{\frac{1}{N_{s}}\sum^{N_{s}}_{i=1} f_{\bm{\theta}(x_i)} }
\end{equation}

where $N$ is the total number of test samples, $N_s$ is the number of test samples in specific strata, $f_{\bm{\theta}}(x_i)$ is the model prediction, and $y_i$ is the reference or validation data.

\subsubsection{Independent assessment of countrywide vegetation height maps}

We evaluate the generated vegetation height maps against the independent validation dataset $VHM_{V}$. We compare the predicted $VHM_{S2}$ with $VHM_{V}$, focusing on forest areas only within the forest mask. As we have observed a few topography-induced erroneous vegetation height values higher than $50~m$ in the $VHM_{V}$ in areas with steep slopes e.g. at the edge of riverbeds, we also mask out these 177 pixels in the mean and 7,654 pixels in the maximum $VHM_{V}$ from the comparison. We then split both mean and maximum $VHM_{V}$ into two annual $VHM_{V}$, each containing only data captured within one of the years 2018 or 2019 (see Figure~\ref{FIG:reference_year_ch}). \YUCHANG{We compute the density scatter plots for both year pairs of VHM.} \YUCHANG{We also calculate the linear models} between the $VHM_{V}$ and $VHM_{S2}$ values in combination with the calculation of the mean estimated $VHM_{S2}$, MBE, MAE, RMSE and the relative MAE (MAEr). The MAEr is defined as MAE divided by the mean estimated $VHM_{S2}$.

Additionally, we calculate these statistical metrics for different strata to perform a thorough analysis for the $VHM_{S2}$ of the years 2018 and 2019. \YUCHANG{We analyze the} following strata on the additional countrywide products of topography, forest mix rate, and tree cover density. \YUCHANG{We use the} DTM to define the topography strata for elevation, slope and aspect in Table \ref{tab:result_strata_validation_2020_terrain}. Three forest mix rate strata are defined based on the product obtained from NFI: 0-24.9\%, 25-74.9\% and 75-100\% broad-leaved trees. Tree cover density information is sourced from the Copernicus High Resolution Layer. Based on this product, the two strata 'open forest' (0-79.9\%) and 'dense forest' (80-100\%) are defined. \YUCHANG{We present an} overview of the defined strata and the number of pixel pairs compared for each stratum in Table \ref{tab:result_strata_validation_2020_terrain} and \ref{tab:result_strata_validation_2020_forest} in Appendix.

\subsubsection{Detection of structural changes in forest}

We analyse the capability of the $VHM_{S2}$ biannual difference to detect the structural changes in the forest of the Canton of Aargau in two ways. First, on pixel level, we compare the biannual $VHM_{S2}$ difference within the forest directly to the $VHM_{V}$ difference using a density scatter plot and the above-mentioned statistics. Second, on object level, we identify those connected pixels with a vegetation height decrease greater than $10~m$ as areas with structural change in the biannual difference of the 1-meter $VHM_{V}$ (before resampling). We filter these areas using a minimum size of 25 $m^{2}$ and define those as change objects, which results in 47,917 objects with area sizes of $25 m^{2}$ - $3.91~ha$. We calculate the mean $VHM_{S2}$ difference within the respective object area for all objects. To estimate the minimal area for change to be detected with $VHM_{S2}$, we then generate box plots of these mean values for different object area ranges and compare them to the value distribution of the $VHM_{S2}$ biannual difference of the unchanged forested area.

\section{Results}

In this section, we first evaluate the overall performance of our model \YUCHANG{against the held-out test area on the reference dataset $VHM_{ALS}$ during the years 2017-2020 in Section \ref{sec:model_eval}}. Then \YUCHANG{in Section \ref{sec:independent_eval}}, we extend the evaluation to the second validation dataset\YUCHANG{ $VHM_V$ covering the year 2018 and 2019}, which enables a finer stratified analysis of the performance, as well as a feasibility study for structural change detection based on our model's prediction. 

\subsection{Vegetation height estimation}\label{sec:model_eval}

\paragraph{\bf Overall performance} We evaluate the trained model on the held-out test areas with the reference data of the years 2017 to 2020. Additionally, we perform an ablation study to examine the impact of including the DTM as input in the model. The results shown in Table \ref{tab:result_als_val_nogt40} include the evaluation of the mean and maximum height for different years in two model settings: with and without DTM. Both models' predictions show an MAE under $2~m$ for mean height and under $2.5~m$ for maximum height. The errors of the model with DTM are slightly smaller. The MBE of both models for mean and maximum vegetation height are near zero. Comparing the errors of mean and maximum height, the errors of maximum height are always higher in all years and models. Errors are noticeably different across these four years though: the RMSE of the mean height for the model with DTM in 2019 is $2.98~m$ while the RMSE in 2017 is $2.22~m$, for example. As the reference data from these years cover different parts of Switzerland, both the temporal and spatial differences may contribute to different errors here.

\begin{table}[th]
\caption{Evaluation results of mean and max vegetation heights with reference data VHM for models with or without DTM as input. We report the Mean Bias Error (MBE), Mean Absolute Error (MAE), and Root Mean Squared Error (RMSE), expressed in meters.}
\centering
\begin{adjustbox}{width=\textwidth}
\begin{tabular}{lcccccccccccccccc}
\toprule
 & with & & \textbf{2017} &  &  & \textbf{2018} &  &  & \textbf{2019} &  &  & \textbf{2020} &  &  & \textbf{Overall} \\  
 &   DTM &MBE&MAE&RMSE & MBE & MAE & RMSE & MBE & MAE & RMSE & MBE & MAE & RMSE & MBE & MAE & RMSE \\ \cmidrule(lr){1-2}\cmidrule(lr){3-5} \cmidrule(lr){6-8} \cmidrule(lr){9-11} \cmidrule(lr){12-14} \cmidrule(lr){15-17}
\textbf{Mean} & \multirow{2}{*}{yes} &  -0.30  & 0.97 & 2.22 & -0.47  & 1.57 & 2.91 & -1.02 & 1.62 & 2.98 & -0.19  & 1.39 & 2.55 & -0.52 & 1.44 & 2.74 \\
\textbf{Max} &  & -0.76 & 1.73 & 3.52 & -0.75 & 2.03 & 3.44 & -1.64 & 2.50 & 4.43 & -0.31 & 1.97 & 3.45 & -0.88 & 2.09 & 3.74 \\\cmidrule(lr){1-2} \cmidrule(lr){3-5} \cmidrule(lr){6-8} \cmidrule(lr){9-11} \cmidrule(lr){12-14} \cmidrule(lr){15-17}
\textbf{Mean} & \multirow{2}{*}{no}& -0.39 & 1.05  & 2.41 & -0.80 & 1.69 &3.11 & -0.66 & 1.52 & 2.81  & -0.38 & 1.42 & 2.62 & -0.59 & 1.47 & 2.81  \\
\textbf{Max} && -1.09 & 1.89 & 3.82 & -1.43 &2.29  & 3.82 & -1.29 & 2.37 & 4.29 & -0.66 & 2.00 & 3.54 & -1.15 & 2.18& 3.89 \\
\bottomrule
\end{tabular}
\end{adjustbox}
\label{tab:result_als_val_nogt40}
\end{table}

\begin{figure}
    \centering
    \includegraphics[width=.9\textwidth]{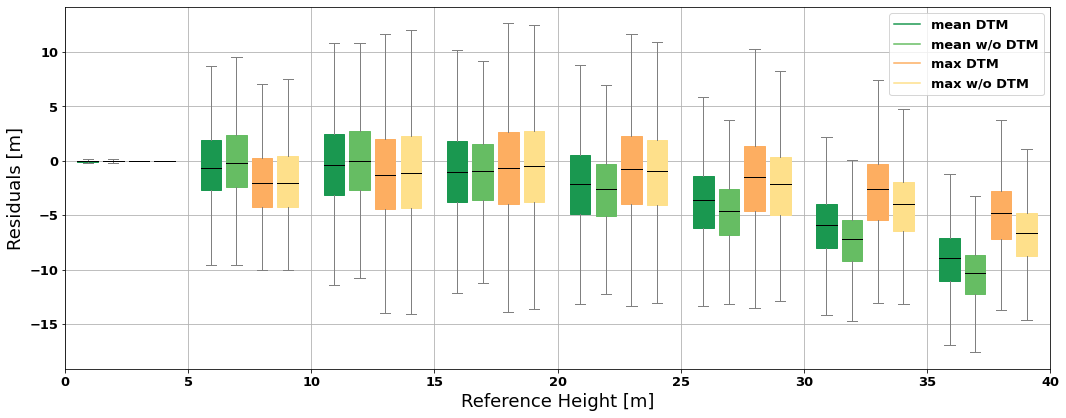}
    \caption{Residuals of models with (\textbf{DTM}) and without DTM (\textbf{w/o DTM}) per $5~m$ height intervals. Note how the performance degrades for high vegetation. In comparison, the model achieves low residuals on the more frequent part of the height distribution (from $5$ to $25~m$).}
    \label{FIG:hist_result_me}
\end{figure}

\paragraph{\bf Performance across the height range} \VIVIEN{As visible on Figure \ref{FIG:hist_result_me} showing the detailed performance of the model, }
residuals can reach 20\% of the reference height when the vegetation height in the reference data is around $40~m$ in both model settings. This indicates a bias towards underestimating high vegetation values. In general, residuals fluctuate around zero in most areas but can become rather large with values larger than $5~m$ in areas with vegetation higher than $30~m$.
Figure \ref{FIG:hist_result_me} also exhibits the positive contribution of using the DTM as additional input.  Indeed, the residuals of the model with DTM are significantly closer to zero on the taller trees, e.g., $\sim 1$m smaller residuals for trees higher than $25$m. Since these taller trees are less frequent, the improvement achieved on this part of the distribution only entails a small improvement on the aggregated metrics of Table \ref{tab:result_als_val_nogt40}. However, Figure \ref{FIG:hist_result_me} clearly shows how the DTM helps mitigating the underestimation problem. 
Therefore, we utilize the model that incorporates the DTM for the subsequent independent assessment of vegetation height maps. 

\begin{figure}[!htbp] 
    \centering
    \begin{tabular}{c|llll}
Input Image &      & \makecell{Prediction} & \makecell{Reference} & \makecell{Error} \\
 \includegraphics[width=.19\linewidth,valign=m]{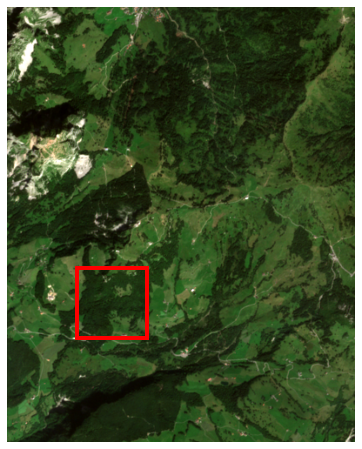}
 & \rotatebox[origin=c]{90}{Mean}  &    
 \includegraphics[width=.22\linewidth,valign=m]{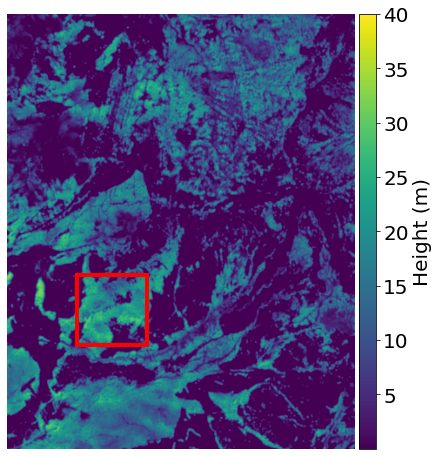}         & 
 \includegraphics[width=.22\linewidth,valign=m]{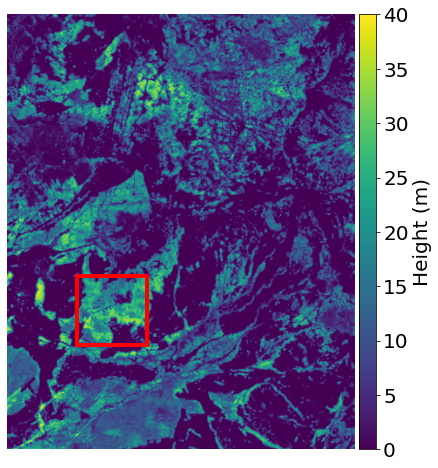} &
 \includegraphics[width=.22\linewidth,valign=m]{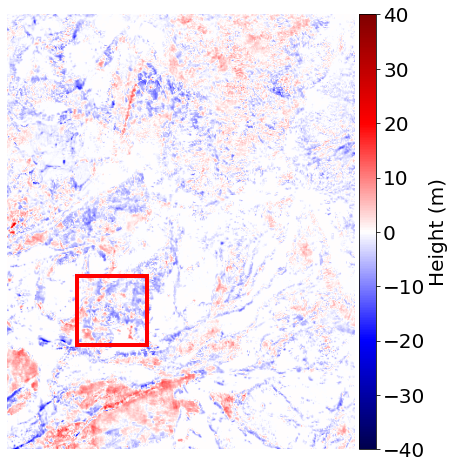}
 \\
 \includegraphics[width=.19\linewidth,valign=m]{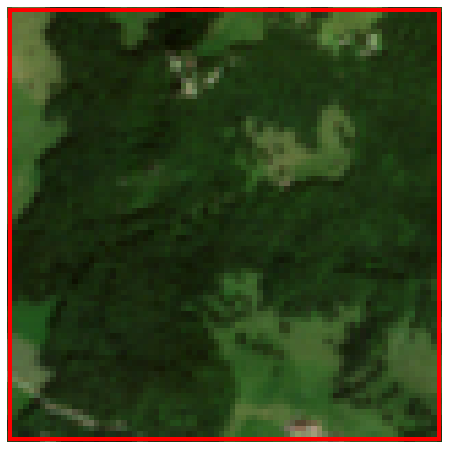}
 & \rotatebox[origin=c]{90}{Mean (zoomed-in)}  &    
 \includegraphics[width=.19\linewidth,valign=m]{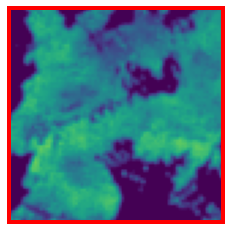}         & 
 \includegraphics[width=.19\linewidth,valign=m]{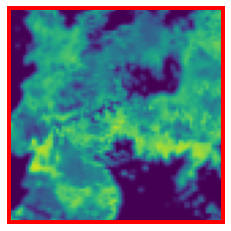} &
 \includegraphics[width=.19\linewidth,valign=m]{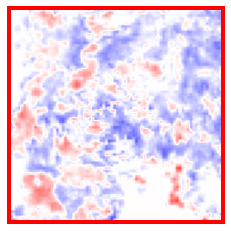}
 \\
& \rotatebox[origin=c]{90}{Max} &   
 \includegraphics[width=.22\linewidth,valign=m]{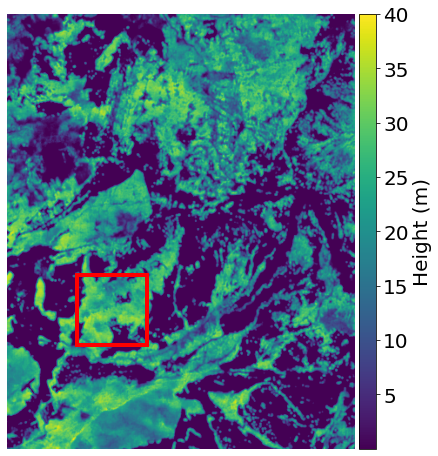}         & 
 \includegraphics[width=.22\linewidth,valign=m]{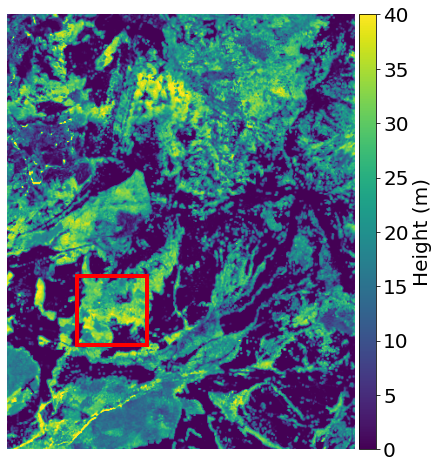} &
 \includegraphics[width=.22\linewidth,valign=m]{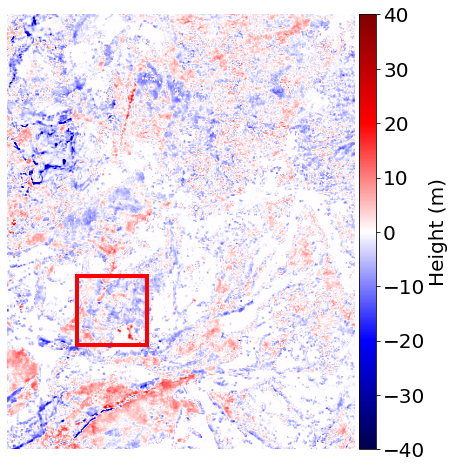}\\
 & \rotatebox[origin=c]{90}{Max (zoomed-in)} &   
 \includegraphics[width=.19\linewidth,valign=m]{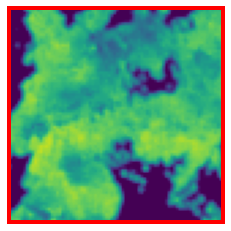}         & 
 \includegraphics[width=.19\linewidth,valign=m]{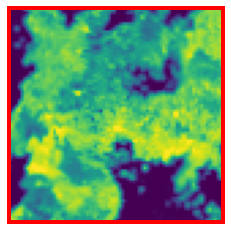} &
 \includegraphics[width=.19\linewidth,valign=m]{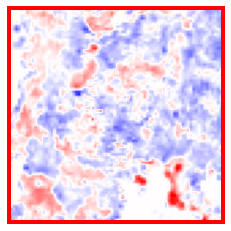}
\end{tabular}
    \caption{Qualitative result of the model with DTM for a $20~km^2$ sample region within the test set in the year 2017. From left to right the columns show the input image, prediction of the model, reference data, and the resulted error (prediction - reference data). The blue color means negative error while the red color means a positive error in the error plot. Note there is an underestimation problem for high vegetation heights, as indicated inside the red rectangle. The zoomed-in figure for the area inside the red rectangle is also shown for better visualisation.}
    \label{fig:result_image_grid}
\end{figure}

\paragraph{\bf Qualitative analysis} \YUCHANG{In Figure \ref{fig:result_image_grid}, we present a qualitative result of a $20~km^2$ sample area from the test set for the year 2017, showcasing the model's successful distinction of various land cover types, such as dense forests and meadows, along with corresponding vegetation height predictions.}
Yet, we can see that finer variations within individual land cover types are not recovered as well by our model's predictions. For example, our model detects dense forests quite well but it tends to underestimate trees higher than 35m. %

\subsection{Independent assessment of vegetation height maps}\label{sec:independent_eval}

\subsubsection{Analysis for the years 2018 and 2019}

\YUCHANG{In the top rows of Table \ref{tab:result_annual_validation}, we present the results of the comparison of $VHM_{S2}$ with the independent validation dataset $VHM_{V}$. These results indicate higher errors in the independent evaluation compared to Table \ref{tab:result_als_val_nogt40}.}
For example, the RMSE values in Table \ref{tab:result_als_val_nogt40} range from $2.22$ to $4.43~m$ while they range from $5.38$ to $5.61~m$ in the top rows of Table \ref{tab:result_annual_validation}. The MBEs are slightly below zero with values between $-0.71$ and $-1.28~m$. 
The differences in the reported metrics between the two different years are similar. Because no data in that area of the year 2018 are used for model training (see Fig. \ref{FIG:reference_year_ch}), this outcome suggests that training on a multi-year dataset leads to good spatial generalization across unseen areas. To further assess temporal generalization, we withhold the training data from the years 2018 and 2019 separately and conduct independent evaluations for each based on the independent $VHM_{V}$. The results of temporal generalization are shown in the bottom rows of Table \ref{tab:result_annual_validation}. We observe that the errors in are slightly larger. Specifically, the overall mean absolute error (MAE) for mean vegetation height increases from 4.30 to $4.87~m$. Moreover, we note that the increase in errors is more prominent in the year 2018 compared to 2019. This can be attributed to the model's need to generalize across both space and time, which presents additional challenges.

\begin{table}[th]
\caption{Result of the comparison of the $VHM_{S2}$ with the independent ALS vegetation height reference. Top rows: full training set results. Bottom rows: temporal generalization with data from 2018 or 2019 withheld. The statistics shown are Mean Bias Error (MBE), Mean Absolute Error (MAE) and Root Mean Squared Error (RMSE), in units meters. }
\centering
\begin{tabular}{lcccccccccc}
\toprule
\multicolumn{1}{l}{} & Trained on &  & \textbf{2018} &  & \multicolumn{1}{l}{} & \multicolumn{1}{c}{\textbf{2019}} &  & \multicolumn{1}{l}{} & \multicolumn{1}{c}{\textbf{Overall}} \\
\multicolumn{1}{l}{} &  full training set & \textit{MBE} & {\color[HTML]{000000} \textit{MAE}} & {\color[HTML]{000000} \textit{RMSE}} & \textit{MBE} & {\color[HTML]{000000} \textit{MAE}} & {\color[HTML]{000000} \textit{RMSE}} & \textit{MBE} & {\color[HTML]{000000} \textit{MAE}} & {\color[HTML]{000000} \textit{RMSE}} \\  \cmidrule(lr){1-2}\cmidrule(lr){3-5}\cmidrule(lr){6-8}\cmidrule(lr){9-11}
\textbf{Mean} & \multirow{2}{*}{yes}  & -1.28 & 4.45 & 5.61 & -0.71 & 4.23 & 5.38 & -0.88 & 4.30 & 5.45 \\
\textbf{Max} & & -0.90 & 4.32 & 5.49 & -0.99 & 4.27 & 5.50 & -0.96 & 4.28 & 5.49 \\\cmidrule(lr){1-2} \cmidrule(lr){3-5}\cmidrule(lr){6-8}\cmidrule(lr){9-11}
  \textbf{Mean} & \multirow{2}{*}{no} & -2.69 & 5.24 & 6.57 & -2.43 & 4.71 & 5.97 & -2.51 & 4.87 & 6.15 \\
\textbf{Max} & & -2.15 & 5.10 & 6.37 & -3.24 & 5.05 & 6.32 & -2.92 & 5.07 & 6.33 \\ \bottomrule

\end{tabular}
\label{tab:result_annual_validation}
\end{table}

\subsubsection{Stratified analysis for the years 2018 and 2019}

\begin{figure}
\centering
\begin{subfigure}[b]{.95\textwidth}
   \includegraphics[width=1\linewidth]{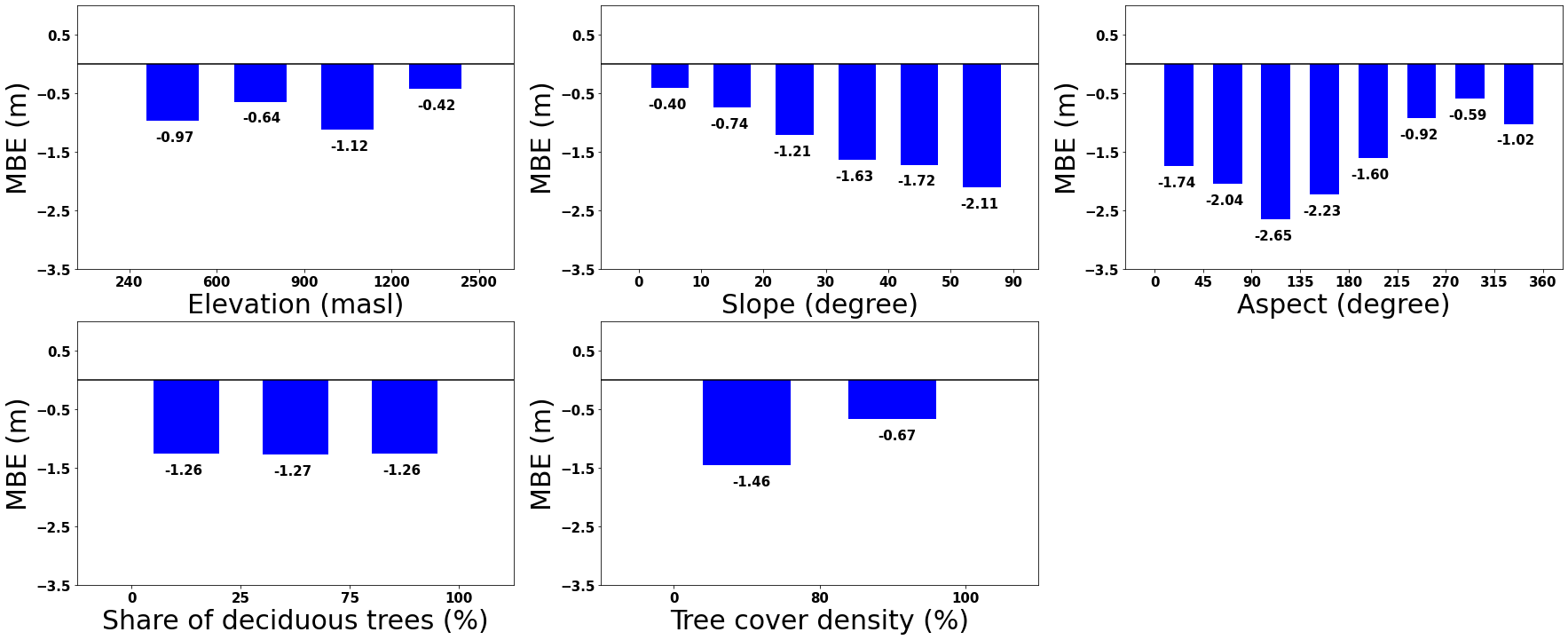}
   \caption{The Mean Bias Error (MBE) for different strata}
   \label{fig:strate_mbe} 
\end{subfigure}
\begin{subfigure}[b]{.95\textwidth}
   \includegraphics[width=1\linewidth]{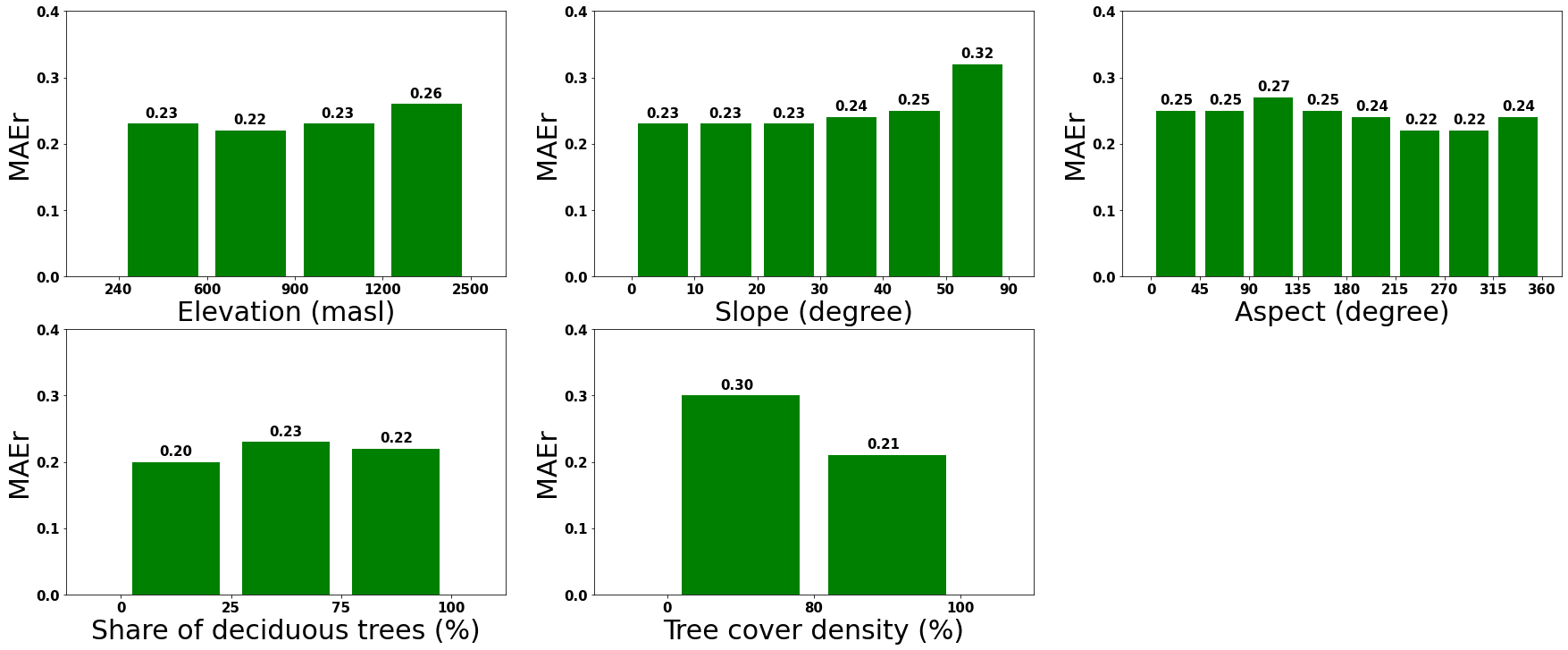}
   \caption{The relative MAE (MAEr) for different strata}
   \label{fig:strate_maer}
\end{subfigure}
\caption{Result of the comparison of the S2 mean VHM with DTM with the independent ALS vegetation height reference within different strata based on \emph{terrain types} and \emph{forest properties}. The Mean Bias Error (MBE) and relative MAE (MAEr) are shown for different strata.}\label{fig:strate_BAR}
\end{figure}

Figure \ref{fig:strate_BAR} shows the result of the stratified analysis of the mean $VHM_{S2}$ with DTM for the years 2018 and 2019. The detailed stratified analysis including the values of all metrics is 
in Table \ref{tab:result_strata_validation_2020_terrain} and \ref{tab:result_strata_validation_2020_forest} in Appendix.
Looking at the elevation strata, higher MAEr values are observed with higher elevation. Both MBE and MAEr values show a negative trend towards steeper terrain in the six slope strata. In the aspect strata, a change from higher MBE in north-west facing slopes (around 300°) to lower ones in south-east facing slopes (around 100°) is observed. The MAEr values show the same trend with a tendency for higher values in south-east facing slopes and lower values in north-west facing slopes.

The underestimation of the vegetation height is equal in forests dominated by coniferous or broad-leaved trees with MBE values of $-1.26~m$ for each. The MAEr values are slightly lower for coniferous forests than for broad-leaved or mixed forests. The two tree cover density strata indicate that underestimation is increased for open forest stands with an MBE of $-1.46~m$ instead of $-0.67~m$. Also, the MAEr is much higher at $0.30$ compared with the denser stands at $0.21$.

\subsection{Detection of structural changes in forest}

\begin{figure}[!htbp] 
	\centering
	    \begin{tabular}{c}
	     $n$ = 4,601,525 \\
		\includegraphics[scale=.4]{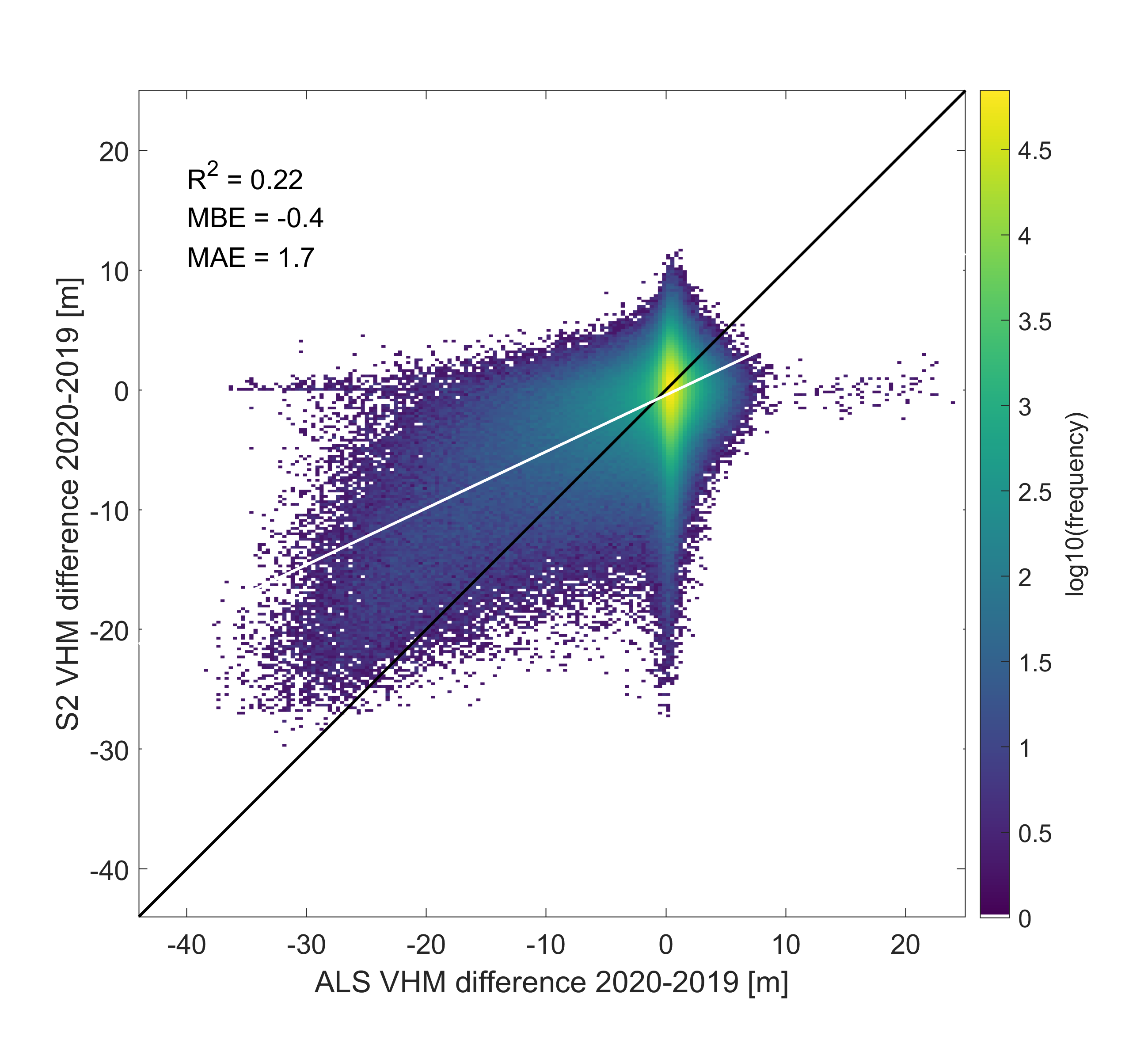}
		\end{tabular}
	\caption{Comparison of the $VHM_{S2}$ and $VHM_{ALS}$ differences (2020-2019) for the forested area of the Canton of Aargau, Switzerland. Mean Bias Error (MBE) and Mean Absolute Error (MAE) in meter are shown. The black line indicates the equivalence line whereas the white one shows the trend line of the derived linear model.}
	\label{fig:densityscatter_S2_ALS_diff}
\end{figure}

\begin{figure}[!htbp] 
	\centering
	
		\includegraphics[scale=.25]{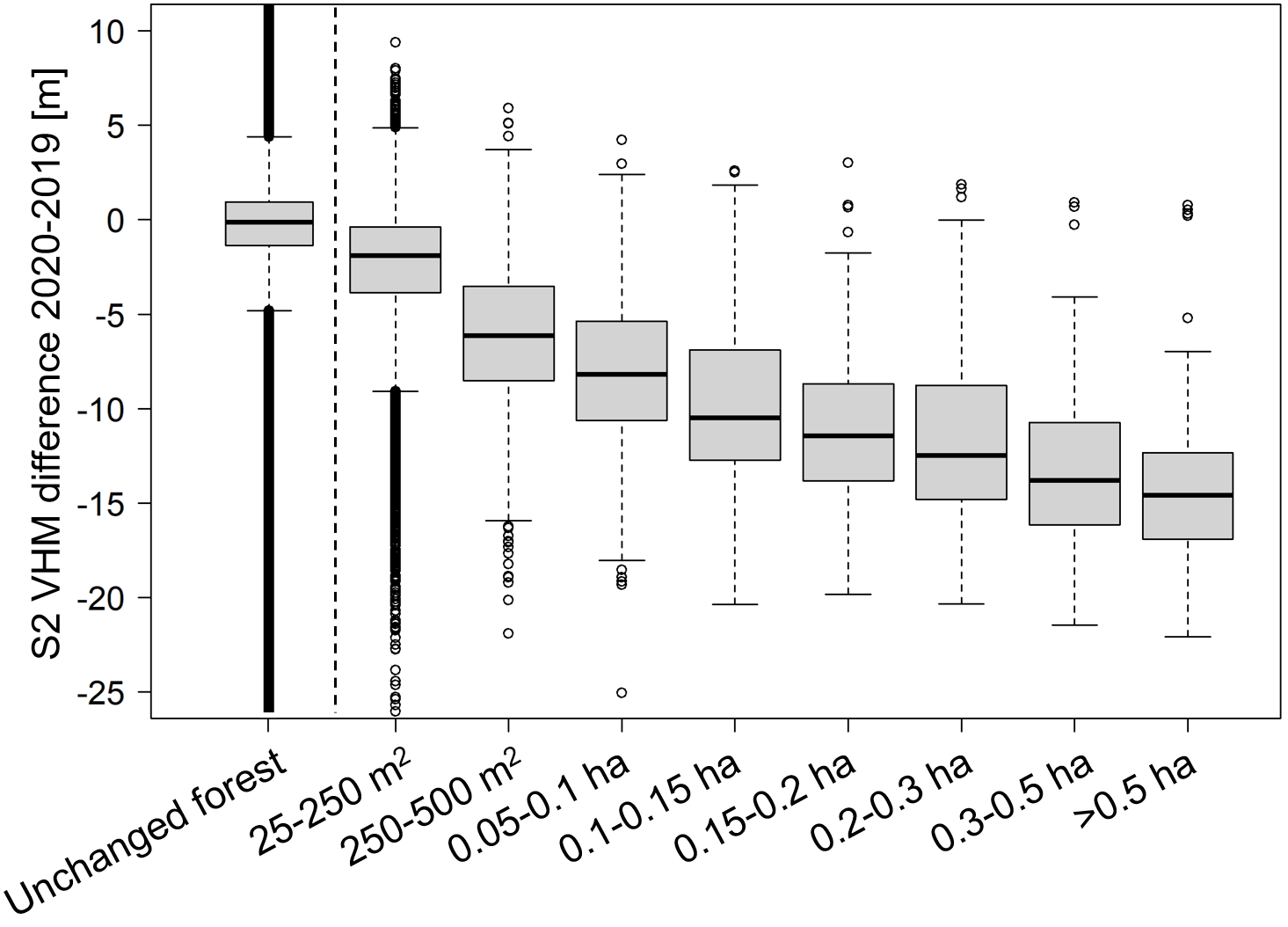}
	\caption{Box plots of the $VHM_{S2}$ difference (2020-2019) within the unchanged forest area of the Canton of Aargau and the mean $VHM_{S2}$ difference in the ALS change objects (vegetation height decrease of $>$10 m) for varying object area size ranges. The boxplots show median, interquartile range (IQR), whiskers (maximum of 1.5 * IQR) and outliers. }
	\label{fig:structural_change_boxplots}
\end{figure}

\YUCHANG{In Figure \ref{fig:densityscatter_S2_ALS_diff}, we present a pixel-level analysis of structural change detection, revealing that the biannual $VHM_{S2}$ difference exhibits a weak relationship with the biannual difference of the ALS data, as indicated by an $R^{2}$ of 0.22.}
The white trend line indicates an underestimation of the height changes of ALS, both in the negative and positive change range. Furthermore, a vertical $VHM_{S2}$ difference value distribution around 0 in the ALS VHM difference is observed indicating pixels containing VH change not matching the ALS reference.
However, the analysis at the object level shows a dependency of the $VHM_{S2}$ mean difference on the size of the change object (Fig. \ref{fig:structural_change_boxplots}). For ALS change objects with a size between $25~m^{2}$ and $250~m^{2}$ the mean $VHM_{S2}$ difference does not differ much compared to unchanged forest. This improves, however, for change objects larger than $250~m^{2}$, where the difference is substantially lower than for unchanged forests. This indicates that most of these objects should be detectable using the $VHM_{S2}$. The larger the change objects, the larger the vegetation height changes. So for change objects larger than $0.1~ha$ and $0.5~ha$, the median values are around -$10~m$ and -$12~m$, respectively. This highlights the capability of $VHM_{S2}$ to effectively capture and detect significant changes, although it may exhibit limitations in accurately detecting small change areas.

\begin{figure}[!htbp] 
	\centering
		\includegraphics[scale=.95]{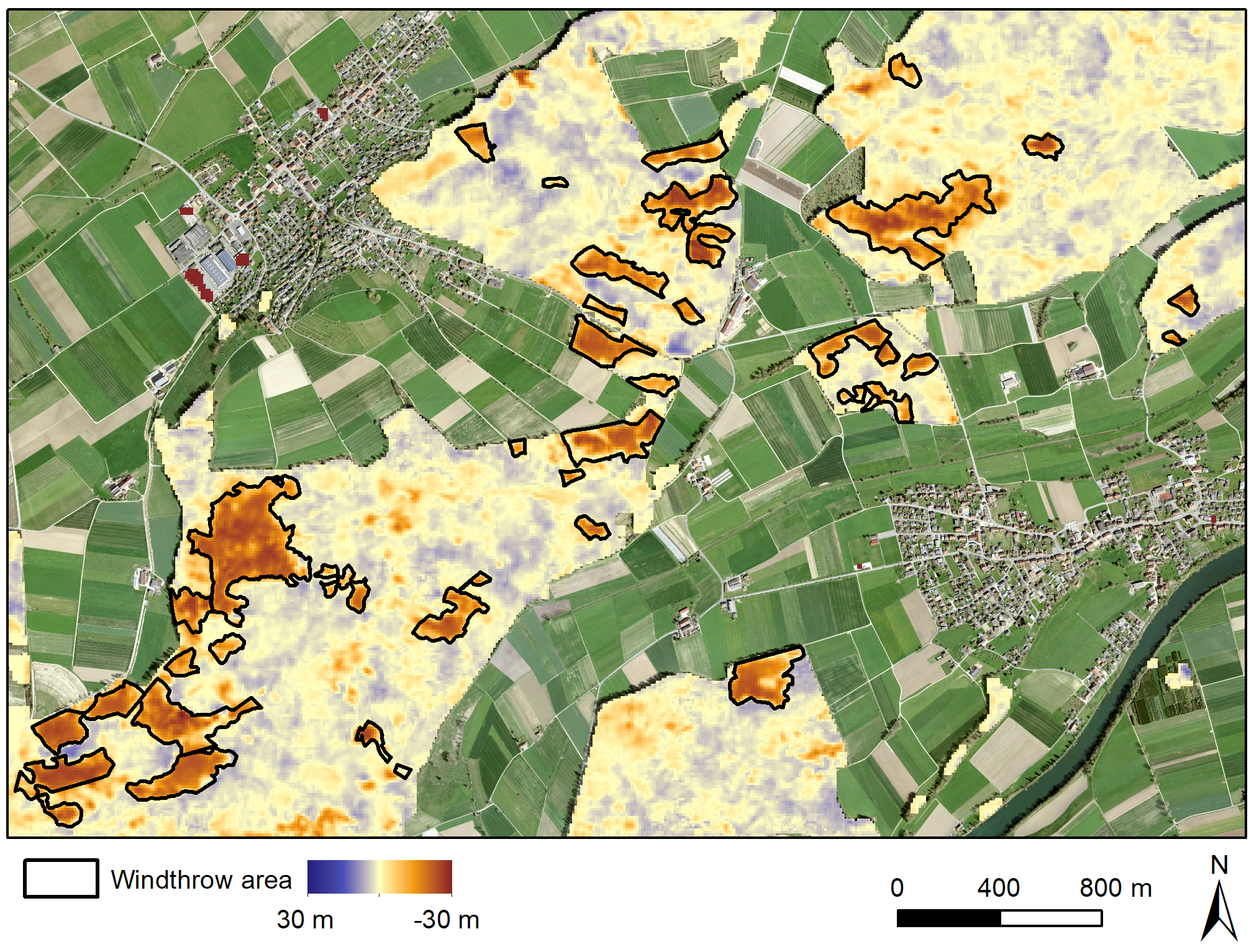}
	\caption{$VHM_{S2}$ difference between 2017 and 2018 within forest. Dark red indicates a strong decrease in vegetation height. The black lines indicate delineated field reference areas for windthrows of the storm "Burglind". The background aerial image is from 2018. © swisstopo}
	\label{fig:windthrow_example}
\end{figure}

While the biannual $VHM_{S2}$ difference primarily captures large changes, it still holds utility for applications such as detecting structural changes caused by windthrows. \YUCHANG{As shown in Figure \ref{fig:windthrow_example}, the structural changes due to the windthrow caused by the winter storm "Burglind" on the 3/1/2018 are well detected by the $VHM_{S2}$ difference between 2017 and 2018.} An ad-hoc change mask defined by pixels with a decrease in vegetation height larger than $10~m$ results in an F1 score of 0.77 for the pixel-level mapping of the field-referenced windthrows. This result indicates the potential of using the produced maps for real applications related to forest dynamics.  

\section{Discussion}

\subsection{Model evaluation}

Our deep learning model achieves an overall MAE $\le 2.5~m$ across different years based on reference data $VHM_{ALS}$. This relatively low error does make VHM regression with modern deep learning and Sentinel-2 images a valuable, complementary tool for countrywide dense mapping at annual repetition rates. %
However, the model tends to underestimate very high vegetation, with the MAE of vegetation heights in the $[30~m, 40~m]$ interval reaching seven meters. \YUCHANG{This saturation towards the high vegetation heights aligns with the previous works \citep{hansen2016mapping, potapov2021landsat+gedi, lang2019country}. }
We argue that there are two main reasons for the underestimation of tall trees. One is the imbalanced VHM reference data used for training. As shown in Figure \ref{FIG:dist_reference_data}, the reference data is dominated by vegetation height lower than $10~m$ and there are only a few samples for vegetation height higher than $30~m$. The model is biased towards vegetation with lower heights, such as bushes or non-vegetated areas. The second reason is the smoothing effect inherent to convolutional neural networks. As discussed in \citet{lang2019country}, a convolutional neural network approach extracts evidence from neighborhoods for a certain pixel, which is prone to smooth out high-frequency information. 
\YUCHANG{Another observation is that max heights are more underestimated than mean heights in short forests, while the reverse holds true for tall forests. This discrepancy can be attributed to the distinct distribution of mean and max vegetation heights. As evident in Figure \ref{FIG:dist_reference_data}, both histograms exhibit another peak, apart from the dominant near-zero values, with mean vegetation heights peaking around $10m$ and max vegetation heights peaking around $25m$. Consequently, the predictions of max height are more prone to underestimation in short forests, as they are far away from the peak of max height, whereas this order is reversed in tall forests, which are closer to the peak of max height.}

\subsection{Independent evaluation of countrywide vegetation height maps}

To refine our analysis, we compare our model’s prediction to an independent vegetation height dataset. The observed errors are larger when compared with the independent $VHM_{V}$. This is mainly due to the reason that only forest areas are considered in this analysis. As Figure \ref{FIG:hist_result_me} shows, the error increases with higher VH typically found in forests. The results also indicate that the inclusion of the DTM does decrease the errors with better MBE and MAE values observed for both years. Therefore, we recommend including it in the model, given the low additional computation cost it represents. Furthermore, our detailed analysis against the independent ALS dataset leads to the following conclusions:

\paragraph{\bf Spatial generalization} Interestingly, the errors of 2018 are in the same range as 2019. This observation demonstrates the model's good spatial generalization capability. The fact that the area covered by $VHM_{V}$ in 2018, which is not included in the model training using $VHM_{ALS}$, still exhibits good performance indicates that the model is able to effectively generalize to unseen spatial regions.

\paragraph{\bf Temporal generalization} Overall, the results of temporal generalization demonstrate that it is feasible to transfer the model to a year without training data while maintaining similar prediction accuracy. Notably, when we exclude the training data from a specific year and conduct independent evaluations, we observe a slight decrease in accuracy. The decrease is relatively larger than the decrease in spatial generalization, suggesting that temporal generalization plays a more dominant role in performance compared to spatial generalization.
The higher errors observed in 2018 compared to 2019 can be attributed to two main factors. Firstly, the model needs to generalize across both spatial and temporal dimensions, which presents a more challenging task. Secondly, the dry summer of 2018 resulted in atypical vegetation behavior, including early wilting, as reported in previous studies \citep{brun20_drought}. This suggests that incorporating meteorological data as additional input to the model, although increasing computational load, could potentially provide benefits in capturing such environmental influences.

\paragraph{\bf Stratified analysis} The strata analysis reveals varying accuracy depending on the topography, forest mix rate or forest density. Two hypotheses can explain the higher MAEr values observed in higher elevations and steep areas. Firstly, the limited availability of training data in these regions could contribute to the increased errors. Secondly, forests in high mountains may be affected by shadows, which can vary depending on the capture time and satellite angle \YUCHANG{\citep{wangchuk2020mapping}}. These factors can introduce additional complexities in estimating vegetation height accurately.
An interesting observation is made concerning the effect of the terrain aspect. The higher underestimation in south-east facing and lower underestimation in north-west facing slopes could be explained by the use of Sentinel-2 data in the L2A processing level. The sen2cor workflow has a tendency to over-correct north-facing slopes, resulting in artificially inflated reflectance values \citep{simonetti2021pan_overestimate}. Replacing the sen2cor by other Sentinel-2 Level 2 processing tools like Force \citep{frantz2019force} may relieve the problem. 
The reason behind the slightly higher accuracy (lower MBE and MAEr) for forests consisting of broad-leaved trees remains unclear. A possible explanation would be a stronger correlation in broad-leaved trees between the diameter of the tree crown and the height. However, this would have to be investigated in more depth. The observed higher MAEr in open forests is not surprising, as open forests usually show a general tendency to lower VHs. Due to the reasons already mentioned above, lower VHs are overestimated by our model.

\subsection{Suitability to detect structural changes in forest}

Our annual VHMs provide possibilities to analyze forest dynamics between two different years. We compare the biannual differences calculated from the generated maps and the independent ALS data and assess the ability of $VHM_{S2}$ to detect structural changes in forests. Our results reveal that it is indeed possible to detect structural changes in forests for two consecutive years if the area of the change is large enough. Smaller changes, like structural changes of individual trees, are, however, not detectable. The relation between the $VHM_{S2}$ and $VHM_{V}$ differences on pixel level implies that changes cannot be reliably detected for individual 10-meter grid cells. 
Successful change detection requires areas $>250~m^{2}$ (Figure \ref{fig:structural_change_boxplots}). %
An inherent reason in our measurement system is the original spatial resolution of the Sentinel-2 data and the smoothing effect introduced by convolutional neural networks. Each individual pixel of the utilized Sentinel-2 channels covers an area of $100~m^{2}$, but due to the convolutional kernels applied during processing, fine textual details are smoothed, leading to a final prediction that may not accurately represent these fine details.
The larger the changed areas become, the less they are impacted by this smoothing effect. Note that this finding also suggests that $VHM_{S2}$ differences detected by comparing two VHMs produced by our model should rather be limited to detect locations and measure the spatial extent of structural change. The current method is not accurate enough to measure the absolute change in vegetation height with reasonable accuracy. Subtle structural changes such as the forest growth between two consecutive years cannot be monitored using the $VHM_{S2}$. Error rates, as shown in Table \ref{tab:result_annual_validation}, are too high to capture the annual forest growth rate of $1~m$ at most. Nevertheless, our results demonstrate successful support for applications such as detecting structural changes caused by windthrows.
\YUCHANG{In addition to windthrow detection, our study's findings hold valuable potential to support the detection of other forest disturbances related to vegetation height changes. In their study, \citet{senf2021mappingForestDisturbance} have successfully generated forest disturbance maps of Europe from 1986 to 2016, revealing that most forest disturbances have a size larger than $0.5~ha$, which is large enough to be detected by our derived vegetation height maps ($>250~m^{2}$). Consequently, our derived results can potentially detect other large-scale forest disturbances as well, consistent with findings by \citet{hansen2013high}. For instance, vegetation height change maps can play a significant role in forest fire monitoring by assessing the extent of damage caused by fires and monitoring the recovery process e.g. every 5 years. As mentioned above, the error rates of the $VHM_{S2}$ prevent it from accurately capturing annual regrowth. The integration of vegetation height change data empowers forest managers to foster resilient forest ecosystems and ensure the sustainable use of forest resources. Overall, our study offers products that can positively impact various forest management activities.}

\section{Conclusion and Outlook}
In this work, we set out to explore the potential of satellite data-derived vegetation height maps as a complementary, faster and more homogeneous alternative to maps based on aerial campaigns. Our results indicate that our $VHM_{S2}$, computed for several consecutive years densely at 10-meter grid spacing for Switzerland, can indeed be put to good use. %
Evaluation of our vegetation height maps on independent, accurate ALS data reveals pros and cons. On the one hand, it demonstrates relatively high mapping accuracy for most parts of the map and also, if predicting for different years, transferability of a trained model between years significantly lowers the computational burden. It also allows for computing longer VHM time series.
On the other hand, our detailed, stratified analysis also reveals that VHM produced with our deep learning approach using Sentinel-2 satellite imagery have biases. Errors are generally higher for tall trees, high altitudes, steep slopes, and differ as a function of the aspect. These findings are very valuable to inform future reference data collection campaigns, both via ALS and in-situ surveys. 
Our results further indicate that VHM of consecutive years produced with our convolutional neural network and individual Sentinel-2 satellite images can successfully detect structural changes $>250~m^2$. %

It is planned to implement the workflow presented in this study within the framework of the Swiss NFI to generate annual, countrywide $VHM_{S2}$ under leaf-on conditions. As soon as more ALS data for the whole of Switzerland is acquired and processed by swissSURFACE3D, the plan is to train the CNN using the ALS reference data with full countrywide coverage. Our hope is that this will support mapping with higher accuracy than our current model.

A further prospect would be the generation of a VHM for the whole Alpine arc using our trained model. The Sentinel-2 based input data is available for the whole area in a similar way as in Switzerland. A challenge will be the spatial transfer of the model. It has to be assessed to what extent the training data from Switzerland are representative enough for the whole Alpine arc. Moreover, it would be interesting to use the temporal information as input directly and let the model benefit from the dynamic processes of forests to improve our model, for example. Another potential direction to improve the model is to adopt Vision Transformers \citep{dosovitskiy2020vit} to avoid smoothness caused by convolutional layers.

\section{Data Availability}
The countrywide vegetation height maps of Switzerland generated for the years 2017-2020 are accessible for download (\url{https://doi.org/10.5281/zenodo.8283347}).

\bibliography{clean_mybibfile}

\newpage

\newpage

\section*{Appendix}

\begin{figure}[!htbp] 
	\centering
		\includegraphics[width=.5\textwidth]{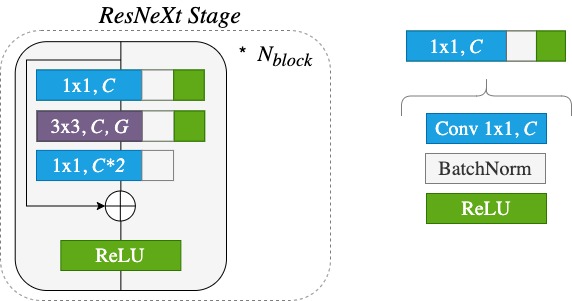}
	\caption{A ResNeXt stage is a stack of $N_{block}$ basic ResNeXt blocks. The basic ResNeXt block consists of three steps: the first one and the third one are made up with a 1x1 convolution layer, batch normalization and ReLU activation, while the second layer replaces the standard convolution with a group convolution \citep{krizhevsky2012groupConv}. The group convolution divides the input and output channels into several groups and performs a separate convolution for each group independently. The aggregated transformation of group convolution achieves improvements compared to ResNet \citep{he2016resnet} while preserving the complexity \citep{xie2017resNeXt}. Inside each ResNeXt block, we use the residual connection to perform identity mapping in order to provide stable gradients. }
	\label{fig:resnext_stage_arch}
\end{figure}

\begin{table}[!htbp] 
\caption{Detailed configuration of the model. We mainly use convolutions with a kernel size of 1 to preserve the spatial dimension. For convolutions with 3 as kernel size in the ResNeXt block, we use zero-padding to keep the spatial dimension. Except this, we use the default setting of the ResNeXt block.}
    \centering
\begin{tabular}{c|c|c} \hline
\textbf{Component}    & \textbf{Layer} & \textbf{Output}    \\ \hline
pixel-level extractor & Conv $1 \times 1$          & {[}2, 15, 15{]}    \\ 
                      & ReLU           & {[}128, 15, 15{]}  \\
                      & Conv $1 \times 1$          & {[}256, 15, 15{]}  \\ \hline
entry block           & Conv $1 \times 1$          & {[}64, 15, 15{]}   \\
                      & BatchNorm      & {[}64, 15, 15{]}   \\
                      & ReLU           & {[}64, 15, 15{]}   \\ \hline
ResNeXt stage 1       & $N_{block}$ = 2      & {[}256, 15, 15{]}  \\
ResNeXt stage 2       & $N_{block}$ = 3        & {[}512, 15, 15{]}  \\
ResNeXt stage 3       & $N_{block}$ = 5        & {[}1024, 15, 15{]} \\
ResNeXt stage 4       & $N_{block}$ = 3       & {[}2048, 15, 15{]} \\ \hline
regression head       & Conv $1 \times 1$       & {[}512, 15, 15{]}  \\
\multicolumn{1}{l|}{} & ReLU           & {[}512, 15, 15{]}  \\
\multicolumn{1}{l|}{} & Conv  $1 \times 1$        & {[}2, 15, 15{]}    \\ \hline
\end{tabular}\label{tab:detailed_model_arch}
\end{table}

\begin{table}[th]
        \caption{Result of the comparison of the S2 mean VHM incl. the DTM with the independent ALS vegetation height reference within different strata based on \textit{\textbf{terrain types}}. The number of pixels in the specific strata $n$, statistics $R^{2}$, mean predicted vegetation height (mean VH), Mean Bias (MBE), Mean Absolute (MAE), Root Mean Square Error (RMSE) and relative MAE (MAEr) are shown for different strata. Mean VH, MBE, MAE and RMSE are all shown in m. The number of analysed pixels varied substantially between the different strata (n). }
    \centering
    \begin{tabular}{ccccccccc}
        Type & Stratum & $n$ & $R^{2}$ & mean VH & MBE & MAE & RMSE & MAEr \\ \hline
        \textbf{Elevation} & \textbf{240-599 masl} & 6,546,426 & 0.61 & 18.30 & -0.97 & 4.29 & 5.46 & 0.23 \\
        ~ & \textbf{600-899 masl} & 3,252,672 & 0.59 & 19.52 & -0.64 & 4.30 & 5.44 & 0.22 \\
        ~ & \textbf{900-1199 masl} & 1,312,755 & 0.59 & 18.60 & -1.12 & 4.35 & 5.48 & 0.23 \\
        ~ & \textbf{1200-2500 masl} & 257,884 & 0.54 & 15.73 & -0.42 & 4.07 & 5.14 & 0.26 \\ \hline
        \textbf{Slope} & \textbf{0-9.9°} & 3,789,662 & 0.62 & 18.05 & -0.40 & 4.06 & 5.19 & 0.23 \\
        ~ & \textbf{10-19.9°} & 3,159,908 & 0.62 & 18.80 & -0.74 & 4.28 & 5.42 & 0.23 \\
        ~ & \textbf{20-29.9°} & 2,664,147 & 0.59 & 19.20 & -1.21 & 4.48 & 5.64 & 0.23 \\
        ~ & \textbf{30-39.9°} & 1,435,093 & 0.56 & 19.00 & -1.63 & 4.56 & 5.73 & 0.24 \\
        ~ & \textbf{40-49.9°} & 248,304 & 0.53 & 17.91 & -1.72 & 4.53 & 5.72 & 0.25 \\
        ~ & \textbf{$\geq$50°} & 72,623 & 0.47 & 14.32 & -2.11 & 4.52 & 6.03 & 0.32 \\ \hline
        \textbf{Aspect} & \textbf{0-44.9°} & 247,058 & 0.53 & 20.54 & -1.74 & 5.10 & 6.33 & 0.25  \\
        \textbf{($\geq$30° Slope)} & \textbf{45-89.9°} & 149,540 & 0.57 & 20.24 & -2.04 & 4.99 & 6.23 & 0.25 \\
        ~ & \textbf{90-134.9°} & 167,024 & 0.56 & 18.62 & -2.65 & 4.94 & 6.23 & 0.27 \\
        ~ & \textbf{135-179.9°} & 410,036 & 0.57 & 16.92 & -2.23 & 4.26 & 5.42 & 0.25 \\
        ~ & \textbf{180-214.9°} & 181,314 & 0.58 & 16.52 & -1.60 & 3.99 & 5.09 & 0.24 \\
        ~ & \textbf{215-269.9°} & 107,478 & 0.56 & 18.10 & -0.92 & 4.01 & 5.07 & 0.22 \\
        ~ & \textbf{270-314.9°} & 156,758 & 0.53 & 19.42 & -0.59 & 4.28 & 5.36 & 0.22 \\
        ~ & \textbf{315-360°} & 336,812 & 0.52 & 19.67 & -1.02 & 4.72 & 5.88 & 0.24 \\ \hline
    \end{tabular}
        \label{tab:result_strata_validation_2020_terrain}
\end{table}

\begin{table}[th]
        \caption{Result of the comparison of the S2 mean VHM incl. the DTM with the independent ALS vegetation height reference within different strata based on \textit{\textbf{forest properties}}. The number of pixels in the specific strata $n$, statistics $R^{2}$, mean predicted vegetation height (mean VH), Mean Bias (MBE), Mean Absolute (MAE), Root Mean Square Error (RMSE) and relative MAE (MAEr) are shown for different strata. Mean VH, MBE, MAE and RMSE are all shown in m. The number of analysed pixels varied substantially between the different strata (n). }
    \centering
    \begin{tabular}{ccccccccc}
     Type & Stratum & $n$ & $R^{2}$ & mean VH & MBE & MAE & RMSE & MAEr \\ \hline
        \textbf{Share deciduous trees} & \textbf{0-24.9\%} & 2,037,614 & 0.44 & 21.42 & -1.26 & 4.24 & 5.37 & 0.20 \\
        ~ & \textbf{25-74.9\%} & 1,924,864 & 0.50 & 18.78 & -1.27 & 4.38 & 5.58 & 0.23 \\
        ~ & \textbf{75-100\%} & 6,160,964 & 0.50 & 19.56 & -1.26 & 4.33 & 5.46 & 0.22 \\ \hline
        \textbf{Tree Cover Density} & \textbf{0-79.9\%} & 3,038,094 & 0.66 & 14.37 & -1.46 & 4.37 & 5.64 & 0.30 \\
        ~ & \textbf{80-100\%} & 8,331,643 & 0.54 & 20.17 & -0.67 & 4.27 & 5.38 & 0.21 \\ \hline
    \end{tabular}
        \label{tab:result_strata_validation_2020_forest}
\end{table}

\begin{figure}[h]
	\centering
		\includegraphics[width=.7\textwidth]{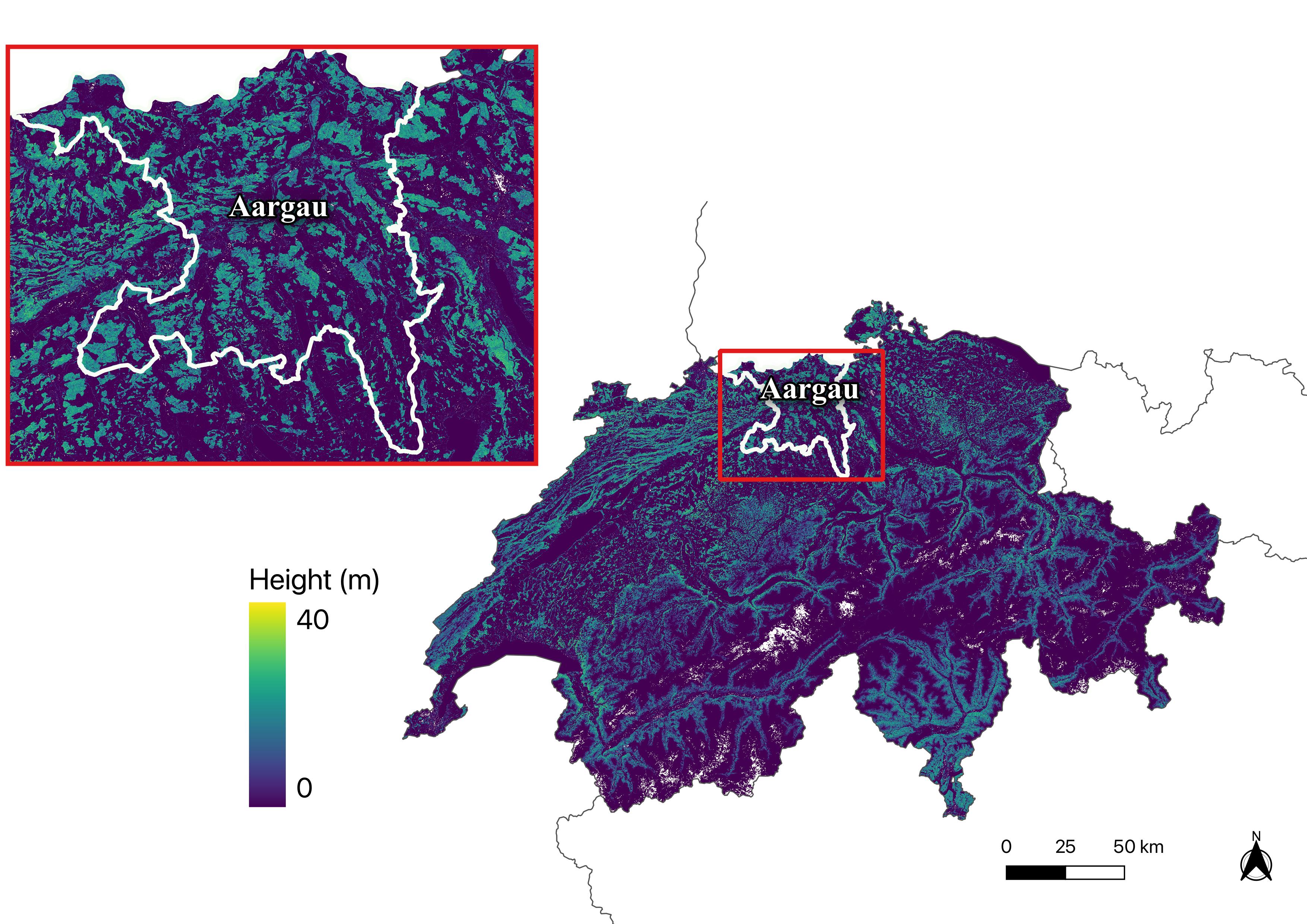}
	\caption{\YUCHANG{The mean vegetation height map of Switzerland of the year 2017 (generated by the model with DTM). The zoomed-in map of canton Aargau is shown in the red rectangle.}}
	\label{FIG:annual_map_2017}
\end{figure}

\begin{figure}
	\centering
		\includegraphics[width=.7\textwidth]{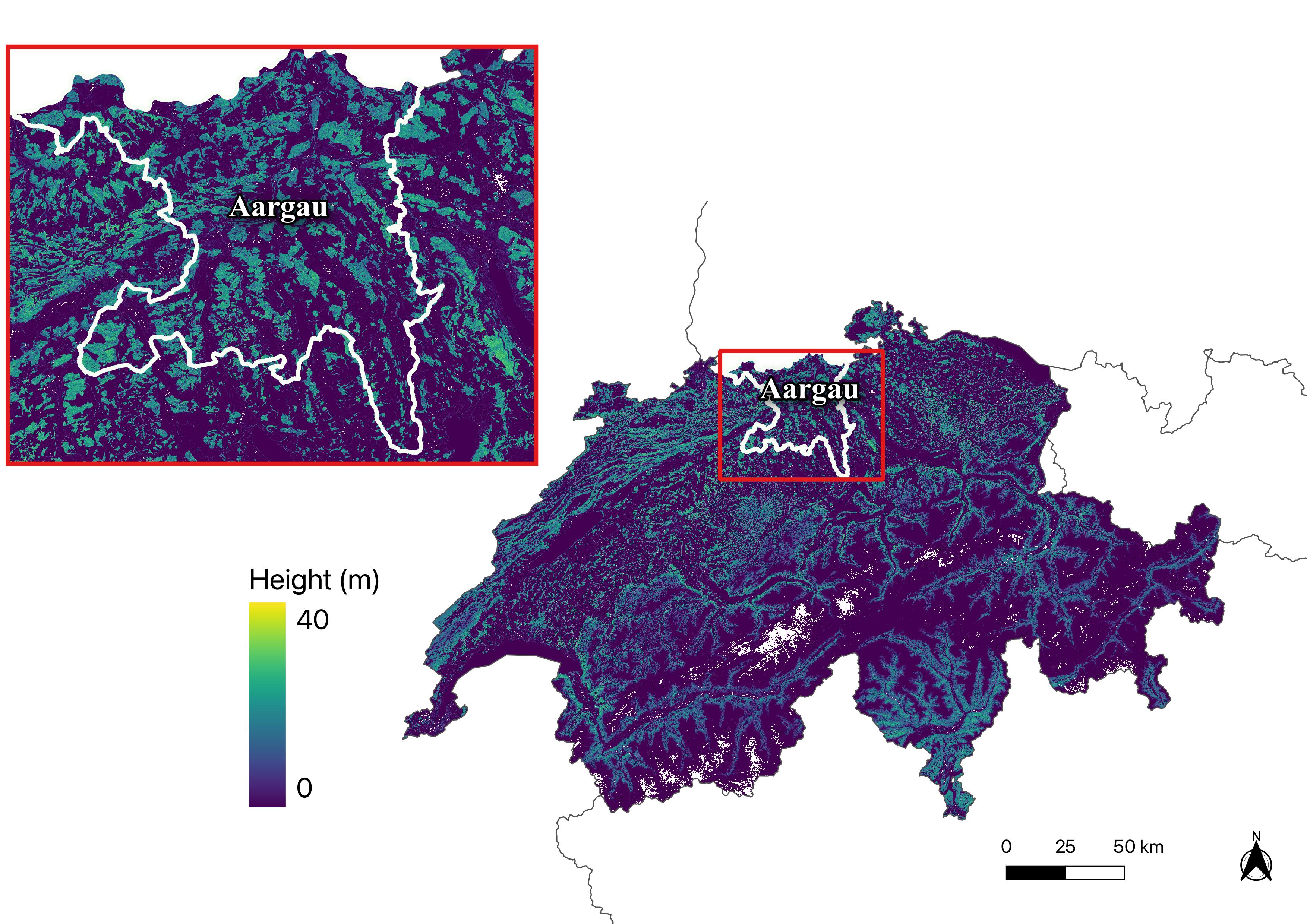}
	\caption{\YUCHANG{The mean vegetation height map of Switzerland of the year 2018 (generated by the model with DTM). The zoomed-in map of canton Aargau is shown in the red rectangle.}}
	\label{FIG:annual_map_2018}
\end{figure}

\begin{figure}
	\centering
		\includegraphics[width=.7\textwidth]{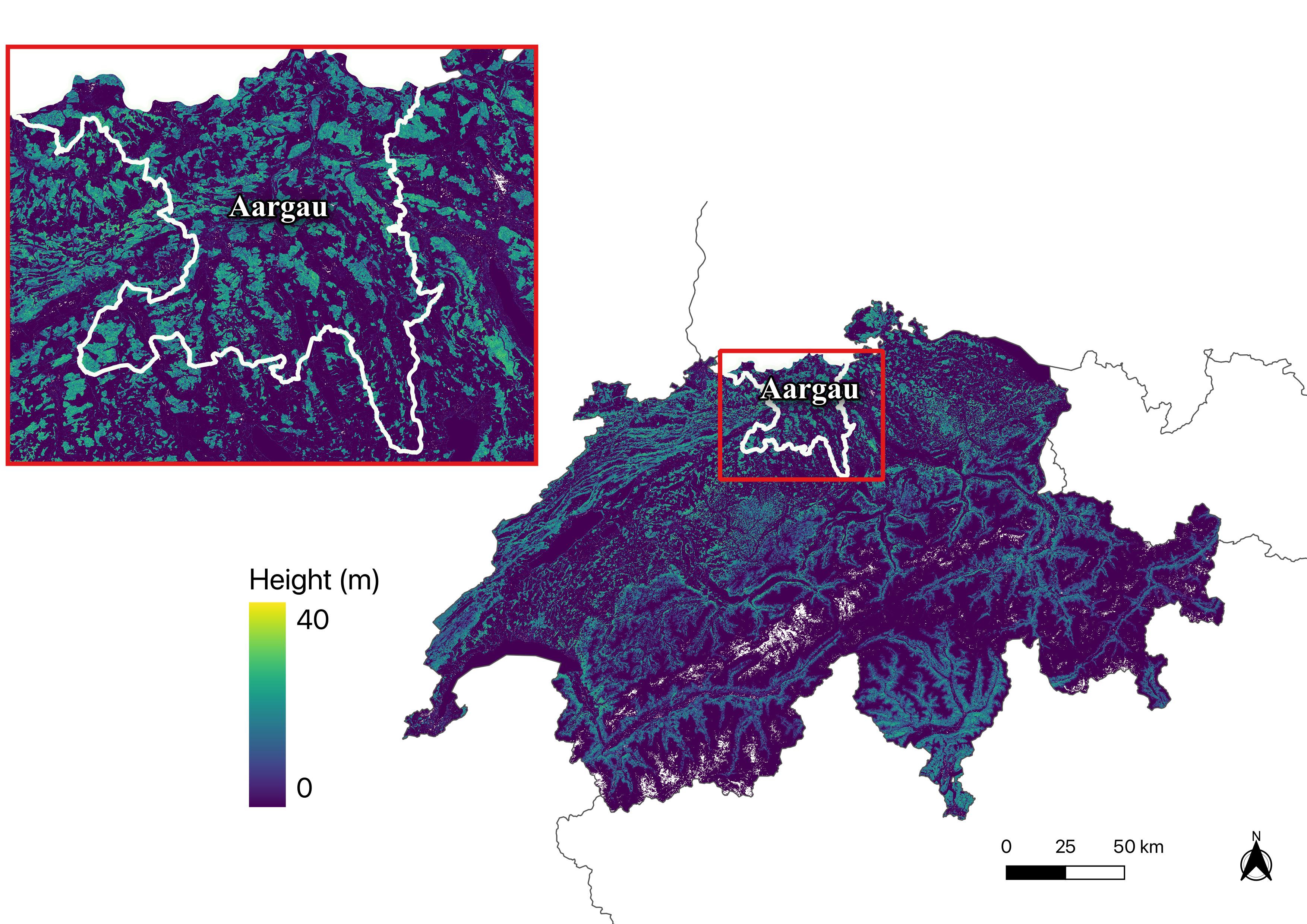}
	\caption{\YUCHANG{The mean vegetation height map of Switzerland of the year 2019 (generated by the model with DTM). The zoomed-in map of canton Aargau is shown in the red rectangle.}}
	\label{FIG:annual_map_2019}
\end{figure}

\begin{figure}
	\centering
		\includegraphics[width=.7\textwidth]{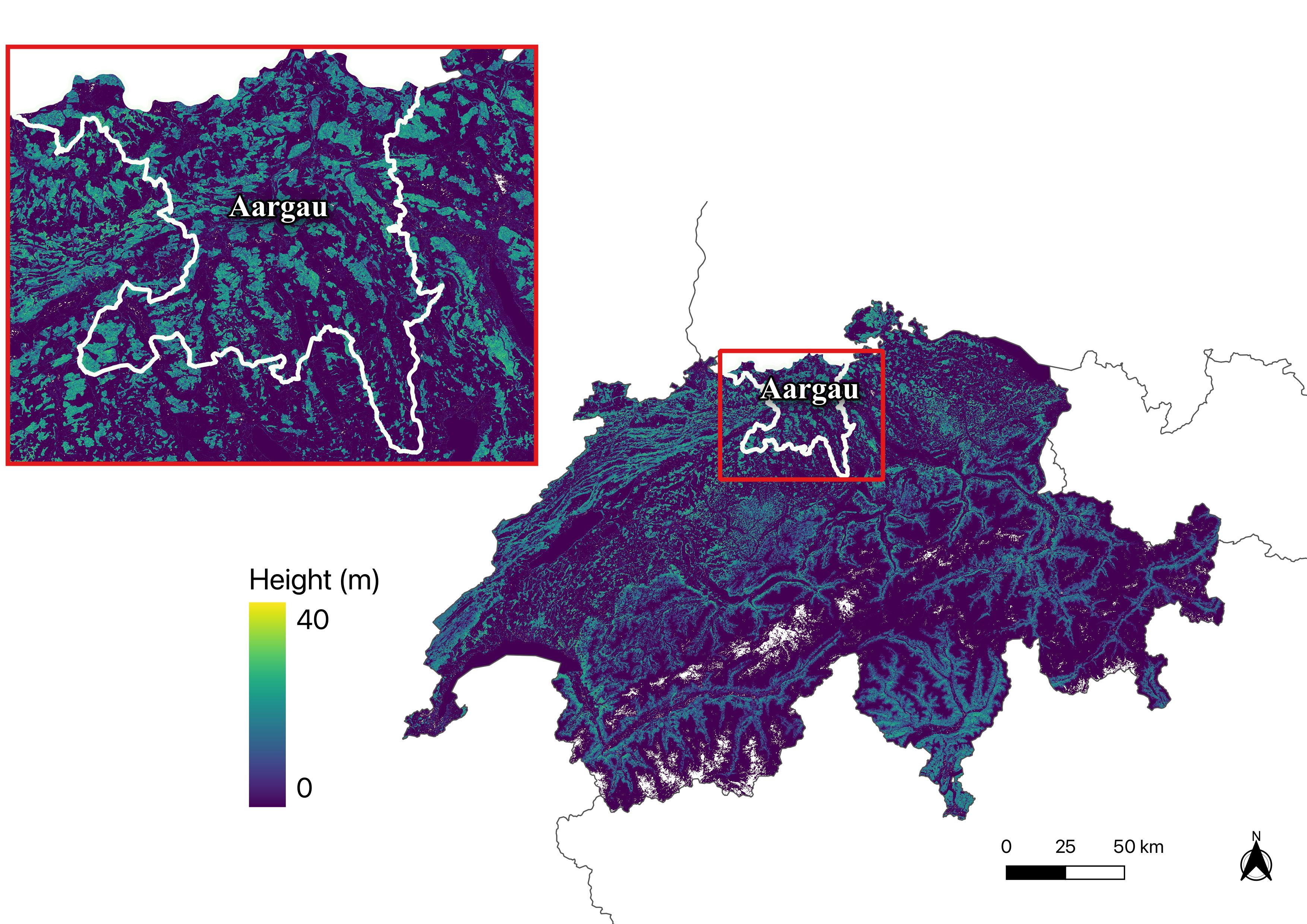}
	\caption{\YUCHANG{The mean vegetation height map of Switzerland of the year 2020 (generated by the model with DTM). The zoomed-in map of canton Aargau is shown in the red rectangle.}}
	\label{FIG:annual_map_2020}
\end{figure}

\end{document}